\documentclass{article}
\usepackage{microtype}
\usepackage{graphicx}
\usepackage{booktabs} 
\usepackage{multirow}
\usepackage{amsmath}
\usepackage{amssymb}
\usepackage{mathtools}
\usepackage{amsthm}
\usepackage{wrapfig}
\usepackage[most]{tcolorbox}
\usepackage{colortbl}
\usepackage{xcolor}
\usepackage{tabularx}
\usepackage{enumitem}

\usepackage{caption}
\usepackage{newtxmath}
\usepackage{subcaption}
\usepackage{algorithm}
\usepackage{algorithmic}
\usepackage{hyperref}

\newcommand{\wrapfill}{\par\ifnum\value{WF@wrappedlines}>0
  \addtocounter{WF@wrappedlines}{-1}%
  \null\vspace{\arabic{WF@wrappedlines}\baselineskip}%
  \WFclear
\fi}

\definecolor{lightgreen}{HTML}{CEEAD6}
\definecolor{lightred}{HTML}{FAD2CF}
\definecolor{lightorange}{HTML}{FEEFC3}
\definecolor{lightblue}{HTML}{30CEFE}
\definecolor{darkerblue}{HTML}{5CA3FF}
\definecolor{brightpurple}{HTML}{9865FE}

\newtcolorbox[auto counter, number within=section]{remarkbox}[1][]{
    colback=olive!5!white,   
    colframe=olive!30!black,   
    fonttitle=\small\itshape, 
    title=Remark~\thetcbcounter: #1,
    boxrule=0.5pt,            
    arc=3mm                   
}

\usepackage[preprint]{corl_2025} 
\usepackage[capitalize,noabbrev]{cleveref}

\renewcommand{\eqref}[1]{(\ref{#1})}
\newcommand{\eg}{\emph{e.g.}}
\newcommand{\ie}{\emph{i.e.}}

\makeatletter
\Crefname{section}{\S\@gobble}{\S\@gobble}
\Crefname{subsection}{\S\@gobble}{\S\@gobble}
\Crefname{proposition}{Prop.}{Props.}
\Crefname{figure}{Fig.}{Figs.}

\renewcommand{\paragraph}{%
  \@startsection{paragraph}{4}{\z@}
                {0.ex}
                {-1em}
                {\normalsize\bf}
}

\makeatother

\title{Fast Flow-based Visuomotor Policies via \\ Conditional Optimal Transport Couplings}

\author{
  Andreas Sochopoulos$^{1,2}$,\ Nikolay Malkin$^1$, \ Nikolaos Tsagkas$^1$,\ Jo\~ao Moura$^1$, \\ 
  \textbf{Michael Gienger$^2$,\ Sethu Vijayakumar$^1$} \\
  $^1$University of Edinburgh \ $^2$Honda Research Institute Europe\\
  \texttt{\href{https://ansocho.github.io/cot-policy/}{ansocho.github.io/cot-policy}}\\
}

\begin{document}
\maketitle

\begin{abstract}     \looseness=-1
     Diffusion and flow matching policies have recently demonstrated remarkable performance in robotic applications by accurately capturing multimodal robot trajectory distributions. However, their computationally expensive inference, due to the numerical integration of an ODE or SDE, limits their applicability as real-time controllers for robots. We introduce a methodology that utilizes conditional Optimal Transport couplings between noise and samples to enforce straight solutions in the flow ODE for robot action generation tasks. We show that naively coupling noise and samples fails in conditional tasks and propose incorporating condition variables into the coupling process to improve few-step performance. The proposed few-step policy achieves a 4\% higher success rate with a 10$\times$ speed-up compared to Diffusion Policy on a diverse set of simulation tasks. Moreover, it produces high-quality and diverse action trajectories within 1--2 steps on a set of real-world robot tasks. Our method also retains the same training complexity as Diffusion Policy and vanilla Flow Matching, in contrast to distillation-based approaches. 
\end{abstract}

\keywords{Flow Matching, Optimal Transport, Imitation Learning}

\section{Introduction}
\label{introduction}

Continuous-time generative models, such as Diffusion Models (DM)~\citep{ho2020denoising} and continous normalizing flows trained by flow matching (FM)~\citep{lipman2022flow}, have recently proven to be an excellent choice for imitation learning of visuomotor robot policies from datasets of expert demonstrations~\citep{urain2024deep, ajay2022conditional, chi2023diffusion}.
This success stems from these models' ability to capture high-dimensional multimodal distributions, in this case, over actions conditioned on observations.
Conversely, explicit models~\citep{mandlekar2021matters, torabi2018behavioral, sochopoulos2024learning} try to learn a direct mapping between robot observations and actions, most often using Behavior Cloning (BC). 

Continuous-time generative models come with a high computational cost for sampling actions, as inference requires numerically solving an ODE~\citep{karras2022elucidating, lipman2022flow} or SDE~\citep{song2020score}. 
The cost of generating a sample grows with the number of time discretization steps used for integration, and the number of steps needed to generate high-quality samples may be large.
ODEs can typically produce accurate actions in fewer integration steps than SDEs, but even 20 steps~\citep{prasad2024consistency} is prohibitive for real-time inference.
Over the past years, there have been significant efforts to reduce the computational cost of inference in continuous-time generative models. 
These efforts have been initiated mostly in the context of image generation~\citep{song2023consistency, yang2024consistency, frans2024one}.

The high latency of action generation using DM/FM policies has a direct impact on their real-world applicability, since it significantly reduces the frequency with which the policy can provide new actions to the robot. This in turn can lead to intermittent robot motions~\citep{chi2023diffusion}, where the robot follows a short-horizon trajectory and then stops to compute the next trajectory, or inability to solve tasks within dynamic environments (\eg, manipulating deformable objects~\citep{ze20243d}).

Recently, interest has also emerged in accelerating robot policies~\citep{prasad2024consistency, wang2024one, hu2024adaflow}. The most common methodology for learning high-quality one-step models is by using distillation techniques that amortize integration of an ODE over a time interval into a single-pass computation~\citep{song2023consistency, yang2024consistency, frans2024one}. However, distillation requires a trained expert or multiple sequential re-trainings of a flow model~\citep{liu2022flow}, significantly increasing the time required to train high-quality one-step models. This can be prohibitive when training large policies on rich datasets with many high-dimensional exteroceptive robot observations. 

In this paper, we present \textit{COT Policy}, a FM methodology for training high-quality few-step visuomotor policies which preserve multimodality without a need for additional training phases.
Our method is based on the use of Optimal Transport (OT) couplings~\citep{tong2023improving} between noise and target actions.
We demonstrate that naively using unconditional OT in FM for training conditional robot policies yields biased flows and is sometimes even worse than vanilla FM.
Instead, we reformulate the OT problem to account for joint distributions of target samples and conditions, similar to recent works in conditional OT (COT) couplings~\citep{kerrigan2024dynamic}.
Our main claims and contributions are the following:

\begin{enumerate}[left=0pt,nosep,label=(\arabic*)]
    \item We introduce COT couplings and a technique for handling continuous conditions by dimensionality reduction and clustering. 
    \item We use COT to train flow-based policies that outperform standard FM and OT-CFM in few-step performance on several robotic tasks.
    \item We analyze the diversity of sampled trajectories and show that COT Policy maintains the multimodality of the action distribution necessary for robust control at low inference cost.
\end{enumerate}

\section{COT Policy}

In this section, we review the preliminaries on flow matching as relevant to our setting and present the proposed approach. Additional background and related work can be found in \Cref{sec:rw}.

\subsection{Preliminaries: Conditional Flow Matching for Generative Modeling}

\paragraph{Neural ODE generative models.} The problem of generative modeling with continuous normalizing flows, or neural ODEs \citep{chen2018neural}, asks to find an ODE that transforms an initial noise distribution $p_0$ over $\mathbb{R}^d$ into a target distribution $p_1$, where samples from both $p_0$ and $p_1$ are available. Given an ODE
\begin{equation}
    dx = v_\theta(t, x)\,dt, 
    \label{eq:neural_ode}
\end{equation}
where $v_\theta:\mathbb{R}\times\mathbb{R}^d\to\mathbb{R}^d$ is a vector field parametrized by a neural network with parameters $\theta$ taking $x$ and $t$ as input, one aims to find $\theta$ such that if $x(0)=x_0\sim p_0$, then the distribution over $x(1)$ induced by integration of the ODE \eqref{eq:neural_ode} from $t=0$ to $t=1$ with initial conditions $x(0)$ matches $p_1$. In our setting, $p_0$ is a fixed Gaussian noise distribution $N(0,I_d)$, so approximate samples from $p_1$ can be generated from the trained model by sampling $x(0)\sim\mathcal N(0,I_d)$ and integrating the ODE (\eg, by \texttt{euler} integration). (The time variable $t$ in the ODE is part of the probabilistic model and is not to be confused with the robot timestep $\tau$, to be introduced later.)

\paragraph{Conditional flow matching.} Conditional flow matching (CFM) assumes the choice of two objects: a \emph{coupling} distribution $q(x_0,x_1)$ whose marginals over $x_0$ and $x_1$ equal $p_0$ and $p_1$, respectively, and a choice of an interpolating path from $x_0$ to $x_1$ for every $(x_0,x_1)$. The latter is an ODE $dx=u(t,x\mid x_0,x_1)\,dt$ whose solution with initial conditions $x(0)=x_0$ has $x(1)=x_1$. In this paper, we always use a linear interpolant: $u(t,x\mid x_0,x_1)=x_1-x_0$, yielding $x(t)=tx_1+(1-t)x_0$, so the solution moves from $x_0$ to $x_1$ along a line segment at a uniform rate as $t$ moves from 0 to 1.

In CFM with linear interpolants, the network $v_\theta$ is trained through the stochastic regression objective:
\begin{align}
    \label{eq:cfm_loss}
    \mathcal{L}_{\text{CFM}}(\theta) := \mathop{\mathbb{E}}_{\substack{t \sim \mathcal{U}([0,1]) \\ x_0,x_1\sim q(x_0,x_1)}} \big\| \underbrace{v_\theta(t, \overbrace{tx_1+(1-t)x_0}^{\text{interpolant}})}_{\text{learned vector field}} - \underbrace{(x_1-x_0)}_{\text{target}} \big\|^2.
\end{align}
The key mathematical fact making this objective correct is that (under smoothness conditions that we always assume to be satisfied), the vector field $v_\theta$ that minimizes the loss \eqref{eq:cfm_loss} among all time-conditional vector fields transforms the initial distribution $p_0$ at $t=0$ into the target distribution $p_1$ at $t=1$, that is, solves the generative modeling problem introduced above.
Notably, training with the objective \eqref{eq:cfm_loss} does not require any ODE integration, making CFM a \emph{simulation-free} algorithm. 

\paragraph{Optimal transport couplings.}

Different choices exist for the coupling $q(x_0,x_1)$. In Independent Coupling CFM (I-CFM), which we use interchangeably with CFM for the rest of the paper, $q(x_0,x_1)=q(x_0)q(x_1)$.  Another option, which was shown in \cite{tong2023improving,pooladian2023multisample} to reduce objective variance and straighten integration curves, takes $q$ to be a minibatch optimal transport (OT) plan computed from batches of samples $x_0,x_1$. Such a coupling approximates the OT plan between $p_0$ and $p_1$. 

We next recall some preliminaries about OT.
The static OT problem aims to find a coupling of minimal cost between two distributions. Given a cost function $C(\cdot, \cdot):\mathbb{R}^d \times\mathbb{R}^d\rightarrow\mathbb{R}$, we search for a solution to the optimization problem:
\begin{align}
    \label{eq:static_ot}
    \mathrm{OT}(p_0,p_1) = \inf_{\pi \in \Pi(p_0,p_1)} \mathbb{E}_{(x_0,x_1)\sim\pi}[ C(x_0,x_1)],
\end{align}
where $\Pi(p_0,p_1)$ denotes the space of probability measures whose left and right marginals equal $p_0$ and $p_1$, respectively. Intuitively, we seek a way to (stochastically) transport particles distributed according to $p_0$ so as to make them distributed according to $p_1$, minimizing the total cost of transportation. The infimum in \eqref{eq:static_ot} is called the OT cost, and a minimizer is called an OT plan.\footnote{Under some conditions, \eg, squared Euclidean cost and $p_0$ absolutely continuous with finite variance, the infinimum is achieved and the minimizer is unique \cite{brenier1991polar}.} In the case where $C(x_0,x_1)=||x_0 - x_1||^2$ is the squared Euclidean distance, the square root of the distance in \eqref{eq:static_ot} is called the 2-Wasserstein distance and denoted $W_2(p_0,p_1)$. 

Under basic conditions, the 2-Wasserstein OT plan exists, is unique up to a null set, and \emph{deterministically} transports $x_0\sim p_0$ to $x_1\sim p_1$. Describing this transport plan as the solution of an ODE, which translates particles from $x_0$ to $x_1$ through time, leads to the dynamic form of the OT problem, also called the Benamou-Brenier form~\citep{benamou2000computational}. We refer to \cite{tong2023improving} for details, but remark that (1) the dynamic OT plan moves points in a straight line segment at uniform rate from $x_0$ to $x_1$; (2) if $q(x_0,x_1)$ is taken to be the 2-Wasserstein OT plan, then the minimizer of the CFM objective \eqref{eq:cfm_loss} among all vector fields is precisely the dynamic OT ODE. 
These properties motivate the use of approximate OT plans as the coupling $q(x_0,x_1)$: the resulting integration paths are more straight and thus the ODE can be solved to suitable precision with fewer \texttt{euler} integration steps.

\paragraph{Conditional generative modeling with CFM.} The CFM framework can be extended to conditional generative modeling. If the distribution $p_1$ varies with a additional variable $c\in\mathcal{C}$, we simply use a coupling $q(x_0,x_1\mid c)$ whose marginals are $p_0$ and $p_1(\cdot\mid c)$ and fit a conditional vector field $v_\theta(x,t\mid c)$, integration of which from $t=0$ to $t=1$ would transform the noise distribution $p_0$ to the conditional target $p_1(\cdot\mid c)$. A conditional form of the static OT problem exists, which aims to minimize transport cost for all values of the condition $c$ simultaneously \cite{hosseini2023conditional}, and \citet{kerrigan2024dynamic} introduced an equivalent Benamou-Brenier dynamic form for the conditional problem, which would be solved by performing CFM with conditional OT couplings $q(x_0,x_1\mid c)$.

The technical challenge that motivates this paper is that when the condition variables $c$ differ for all samples in a dataset, OT couplings are no longer possible to construct, since we cannot construct batches of samples that share the same condition $c$ and use them to approximate the OT plan from $p_0$ to $p_1(\cdot\mid c)$; see \Cref{sec:conditional_ot_couplings}.

\paragraph{Imitation learning as conditional generative modeling.}

As will be described in detail in \Cref{sec:cfm_as_robot_policy}, we treat the learning of actions given expert demonstrations as a conditional generative modeling problem. Assume a dataset of observation-action sequence pairs $\mathcal{D}=\{(o^{(i)},a^{(i)})\}_{i=1}^n$ is given. We treat the $a^{(i)}$ as samples from a target distribution $p_1(a\mid o)$ with conditions $o=o^{(i)}$. Approximating this distribution by learning a neural ODE, conditioned on $o$, that transforms $p_0$ to $p_1(\cdot\mid o)$, would allow sampling actions $a$ given observations $o$ not seen in training.

In our setting, $o$ and $a$ represent observation histories and action sequences over short time horizons, respectively. The dataset $\mathcal{D}$ of short-horizon sequences is formed by extracting short sequences from expert demonstrations of longer duration.
The observations consist of raw RGB images and proprioception~\citep{chi2023diffusion, song2023consistency} but could also include point clouds~\citep{chisari2024learning} or force-torque data~\citep{hou2024adaptive, aburub2024learning}.

\subsection{Optimal Transport Couplings for Finite Conditional Tasks}
\label{sec:conditional_ot_couplings}

\begin{wrapfigure}{R}{.45\textwidth}
    \vspace*{-2em}
\includegraphics[width=\linewidth]{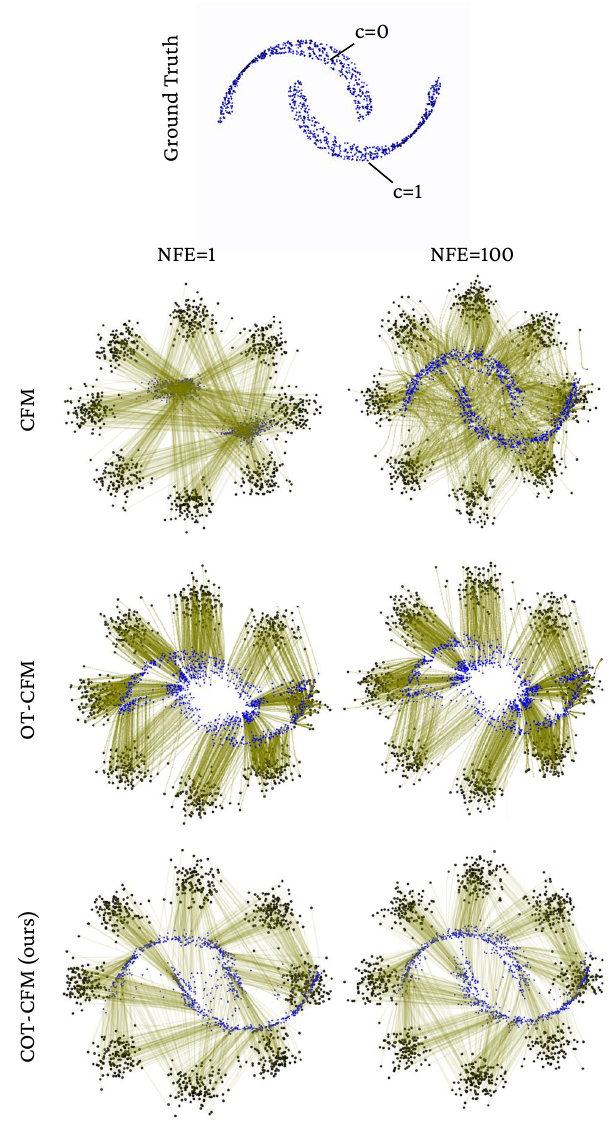}
\caption{(I-)CFM, OT-CFM, and COT-CFM flows trained to generate the two moons distribution from the 8 Gaussians distribution. Generation with 100 (top row) and 1 (bottom row) \texttt{euler} integration steps is shown. CFM gives curved flows that cannot be integrated accurately in one step, while OT-CFM gives biased samples. Our proposed COT-CFM avoids both issues.}
\vspace*{-2em}
\label{fig:moons}
\end{wrapfigure}

For unconditional tasks, taking the coupling $q$ in CFM to be the OT plan between $p_0$ and $p_1$ results in a flow that solves the dynamic OT problem. However, if we are aiming to sample a conditional distribution $p_1(\cdot\mid c)$, then naively taking $q$ to be the OT plan between the noise distribution $p_0$ and $p_1$'s marginal distribution over $x_1$ would not yield a solution to the conditional generative modeling problem, as can be seen in \Cref{fig:moons}, where we train an ODE that transforms $p_0$ (mixture of Gaussians) to $p_1(\cdot\mid c)$ (two moons, where $c=0,1$ indexes the two halves of the moon). Such a naive use of OT-CFM causes noise samples from the top Gaussians to pair almost exclusively with targets from the top moon. During inference, while the paths remain nearly straight, sampling noise from one of the upper Gaussians with $c=1$ leads to out-of-distribution targets. This occurs because few regression targets were established during training to connect the bottom moon to the upper Gaussians.

We first consider conditional tasks with condition variables (or observations) $c\in \mathcal{C}$, where $\mathcal{C}=\{c_1,c_2\dots,c_K\}$ is a finite set. Given a batch of $N$ samples with conditions $B_1=\{(x_1^{(1)},c^{(1)}),(x_1^{(2)},c^{(2)}),\dots,(x_1^{(N)},c^{(N)})\}$, where $x_1^{(i)}\in\mathbb{R}^d$ and $c^{(i)}\in \mathcal{C}$, from the dataset, we construct a batch of noise observations $B_0=\{(x_0^{(1)},c_0^{(1)}),(x_0^{(2)},c_0^{(2)}),\dots,(x_0^{(N)},c_0^{(N)})\}$, where the $x_0^{(i)}$ are independent samples from $p_0$ and $c_0^{(1)},\dots,c_0^{(N)}$ is a uniform-random permutation of $c^{(1)},\dots,c^{(N)}$. 
We then compute the empirical OT map between the empirical sets of sample-condition tuples $B_0$ and $B_1$, using the cost function
\begin{equation}\label{eq:cond_cost}
    c((x_0,c_0),(x_1,c_1))=\|x_0-x_1\|^2+\|\gamma(c_0-c_1)\|^2
\end{equation}
where $\gamma$ is a hyperparameter. This results in a empirical coupling $\pi(x_0,x_1)$, which is used for training the conditional vector field $v_\theta(t,x\mid c)$ conditioned on $c^{(i)}$ using the CFM loss:
\begin{equation}\label{eq:cotcfm}
    \mathcal{L}_{\text{CFM}}(\theta) := \mathop{\mathbb{E}}_{\substack{t \sim \mathcal{U}([0,1]) \\ (x_1^{(i)},c^{(i)})\sim\mathcal{U}(B_1)\\(x_0,c_0)\sim \pi(x_0\mid x_1^{(i)})}} \big\| \underbrace{v_\theta(t, \overbrace{tx_1^{(i)}+(1-t)x_0}^{\text{interpolant}}\mid \overbrace{c^{(i)}\vphantom{)}}^{\text{condition\vphantom{p}}})}_{\text{learned conditional vector field}} - \underbrace{(x_1^{(i)}-x_0)}_{\text{target}} \big\|^2.
\end{equation}
The $c_0$ are only used in computing the OT plan, and taking the $c_0$ to be a permutation of the observed conditions $c_1$ ensures that the two batches of conditions come from the same distribution. We term CFM augmented with this COT pairing as \emph{COT-CFM} (conditional OT conditional flow matching).

The purpose of $\gamma$ is to prioritize pairing of tuples with similar conditions. In the limit of $\gamma\to\infty$, the second term in \eqref{eq:cond_cost} dominates over the first: samples $(x_0,c_0)$ are paired with samples $(x_1,c_1)$ with $c_0=c_1$, and for each given $c$, $\pi$ is the OT plan between the batches of $x_0$ and $x_1$ having condition $c$. On the other hand, if $\gamma=0$, we recover OT-CFM, which ignores the conditions when computing the coupling even while the vector field takes $c$ as an input, leading to biased conditional generation.
We can thus control the trade-off between an exact solution to conditional dynamic OT ($\gamma\to\infty$) and smoothness and generalization ($\gamma\to0$)~\citep{kerrigan2024dynamic}.
An investigation of the effect of $\gamma$ on the solutions of minibatch OT can be found in \Cref{sec:effect_of_gamma}.

The COT-CFM algorithm is applicable even when the set of conditions $\mathcal{C}$ is not finite. However, as we explain next, when the conditions lie in high-dimensional continuous spaces, they will undergo dimensionality reduction and quantization before the OT is computed.

\subsection{COT-CFM as a Robot Policy}\label{sec:cfm_as_robot_policy}

\paragraph{Continuous condition tasks.} In robotic tasks, the conditions for computing robot actions are derived from the robot's sensor observations. 
In our case, these observations $o$ include RGB images and proprioceptive data, both of which take continuous values.

Intuitively, the conditional distribution of robot actions is not sensitive to small changes in the state of the environment and the robot. 
Small deviations in the robot's position or the workspace it operates in do not influence the plan of actions. 
Therefore, we can treat similar states or observations as a single condition in the calculation of $\pi(x_0,x_1)$. 
Effectively, we can obtain an approximation of the continuous conditional OT plan by discretizing the observations and applying the approximation to the conditional OT pairing described in \Cref{sec:conditional_ot_couplings}. This involves the operations:
$e=\mathcal{E}(o),\ \ c=\mathcal{Q}(e)$, 
where $\mathcal{E}$ is an encoder for the observations and $\mathcal{Q}$ is a discretization function. 
The encoder is solely dependent on the modality of the observation, while the discretization function can be any suitable discretization scheme, such as a quantizer~\citep{tian2024visual, zheng2024prise} or a clustering algorithm. 

\begin{wrapfigure}{R}{.53\textwidth}
\begin{minipage}{\linewidth}
    \vspace{-20pt}
    \begin{algorithm}[H]
       \caption{COT Policy training}
       \label{alg:cot_policy}
    \begin{algorithmic}
       \STATE {\bfseries Input:} Noise distribution $p_0$, data-conditions dataset $\mathcal{D}$, condition weight $\gamma$, initial network $v_\theta$, encoder $\mathcal{E}$, discretizer $\mathcal{Q}$.
       \WHILE{ training}
       \STATE Sample data batch $\{(a^{(i)}, o^{(i)})\}_{i=1}^{N} \sim  \mathcal{D}$, noise batch $\{x_0^{(i)}\}_{i=1}^N \sim  p_0$,  $\{t^{(i)}\}_{i=1}^N \sim \mathcal{U}(0, 1)$
       \STATE \textbf{// Noise-action pairing with COT}
       \STATE $e^{(i)} \leftarrow \mathcal{E}(o^{(i)})$, $c_a^{(i)}\leftarrow\mathcal{Q}(e^{(i)})$ 
       \STATE $c_{x_0}^{(1,\dots,N)} \leftarrow {\rm randperm}(c_a^{(1,\dots,N)})$
       \STATE $\pi\leftarrow$ OT plan with cost \eqref{eq:cond_cost} \\\quad from $\{(x_0^{(i)},c_{x_0}^{(i)})\}_{i=1}^N$ to $\{(a^{(i)},c_a^{(i)})\}_{i=1}^N$ 
       \STATE $x_0^{(i)}\sim\pi(\cdot\mid x_1^{(i)})$ \COMMENT{\textcolor{red}{paired initial point}}
       \STATE \textbf{// CFM loss and update}
       
       \STATE $x_t^{(i)} \leftarrow t^{(i)}a^{(i)} + (1-t^{(i)}){x_0^{(i)}}$ \COMMENT{\textcolor{red}{interpolant}}
       \STATE Update $\theta$ with loss \eqref{eq:cotcfm}:\\$\quad\sum_i\|v_\theta(t^{(i)}, x_t^{(i)}\mid o^{(i)})  - (a^{(i)} - x_0^{(i)})\|^2$
        \ENDWHILE
    \end{algorithmic}
    \end{algorithm}
    \vspace{-20pt}
    \end{minipage}
\end{wrapfigure}
 
\paragraph{Robot policies.} Our pipeline for learning a generative policy includes encoding raw RGB observations using a vision encoder that is trained end-to-end with the policy, as is often done in imitation learning using generative models~\citep{chi2023diffusion}. 
We implement $\mathcal{E}$ as PCA on each image batch, leaving the proprioceptive vector unchanged. For the discretizer $\mathcal{Q}$, we perform K-means clustering into $K$ clusters and represent each point by the centroid of the cluster it belongs to. The resulting representations $c=\mathcal{Q}(\mathcal{E}(o))$ are used to compute the OT coupling, while the unprocessed observations $o$ are used for conditioning the flow $v_\theta(t, x \mid o)$ (see \Cref{sec:appendix-training-details} and \Cref{sec:appendix-visual-encoder} for details of the architecture of the model $v_\theta(t,x\mid o)$). This choice of $\mathcal{E}$ and $\mathcal{Q}$, although simple, is easy to extend to any observation modality, adds negligible computational overhead, and avoids the training of additional networks. The pipeline for our policy, called COT Policy, can be seen in \Cref{alg:cot_policy}.

\paragraph{Training complexity.} In contrast to distillation methods, which require an increased number of forward model passes~\citep{frans2024one} or two separate training phases~\citep{prasad2024consistency, wang2024one, lu2024manicm}, COT policies do not require extra training time. Although they include extra processing steps compared to their CFM counterparts, \ie, performing dimensionality reduction and clustering, they add negligible computation time if implemented in a GPU-accelerated framework.

\section{Experiments}
\label{experiments}

We evaluate COT Policy's action-distribution learning capabilities and ability to capture multimodality. Details of demonstration datasets, and implementation can be found in \Cref{sec:implementation} and \Cref{sec:implementation_real}.

\paragraph{Baselines.} COT Policy is compared against \textbf{1)} CFM, \textbf{2)} OT-CFM~\citep{tong2023improving}, \textbf{3)} Diffusion Policy (DP)~\citep{chi2023diffusion}, and \textbf{4)} Adaflow~\citep{hu2024adaflow}, a variance-based adaptive CFM sampler. These choices allow us to support our claim that COT Policy outperforms ODE-based methods in few-step generation without compromising action diversity and with comparable training cost. Although Adaflow requires two-phase training, the second phase trains only linear layers and therefore adds negligible training overhead. Diffusion Policy with a large number of function evaluations per step (NFE) is used for comparisons as a state-of-the-art policy. For all results, unless stated otherwise, COT Policy is implemented with 64 clusters for discretization of the conditions (\Cref{sec:cfm_as_robot_policy}). See \Cref{sec:appendix-training-details} for all implementation details.

\begin{figure}[t]
    \centering
    \includegraphics[width=\linewidth]{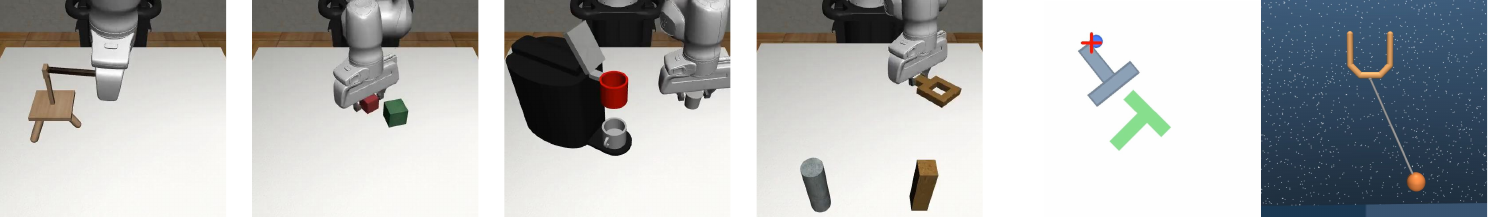}
    \caption{Tasks used for the evaluation of COT Policy.}
    \label{fig:tasks}
\end{figure}

\paragraph{Tasks.} To validate the proposed COT approach, we first consider two low-dimensional synthetic tasks with discrete and continuous conditions (\Cref{fig:moons}). Because the focus of this paper is imitation learning for robot policies, this study is relegated to the Appendix (\Cref{sec:experiments_synthetic}).

In the main text, we consider three sets of simulation tasks: \textbf{1)} four robot manipulation tasks (\Cref{fig:tasks}) from \textit{MimicGen}~\citep{mandlekar2023mimicgen}, which is a RoboSuite~\citep{zhu2020robosuite} and RoboMimic~\citep{mandlekar2021matters} extension, with realistic manipulation environments and the ability to easily generate demonstrations (\Cref{sec:experiments_robotsim}), \textbf{2)} two toy decision making tasks, namely, \texttt{push-t} (commonly used in Behavior Cloning~\citep{florence2022implicit, chi2023diffusion}) and \texttt{cup} from \textit{dm-control}~\citep{tunyasuvunakool2020dm_control} (\Cref{sec:experiments_robotsim}), and \textbf{3)} two Maze navigation tasks from D4RL~\citep{fu2020d4rl} (\Cref{sec:experiments_multimodality}). 

Furthermore, we evaluate our policy on real hardware in three manipulation tasks (\Cref{sec:real_robots}).

\begin{table}[t!]
   \centering
    \resizebox{\linewidth}{!}{
    \begin{tabular}{@{}l c|c c c c c c|c}
        \toprule
        \textbf{Method} & \textbf{NFE} & \texttt{threading\_d0} & \texttt{stack\_d1} & \texttt{coffee\_d1} & \texttt{square\_d0} & \texttt{push-t} & \texttt{cup} & \textbf{Avg} \\
        \midrule
         \footnotesize CFM (\citet{tong2023improving})  & 4 & 0.730 & 0.927 & 0.720 & \textbf{0.753} &  {0.877} & 0.776 & \underline{0.797} \\
         \footnotesize OT-CFM (\citet{tong2023improving}) & 4 & 0.723 & \textbf{0.940} & 0.623 & 0.697 & 0.712 & 0.743 & 0.740 \\
         \footnotesize DP (DDIM) (\citet{chi2023diffusion}) & 4 & 0.720 & 0.917 & 0.650 & 0.650 & 0.877 & 0.763 & 0.763 \\
         \footnotesize DP (DDIM) (\citet{chi2023diffusion}) & 20 & 0.720 & \textbf{0.940} & 0.650 & \underline{0.707} & \textbf{0.881} & 0.787 & 0.781 \\
        \midrule
         \footnotesize Adaflow (\citet{hu2024adaflow}) & 1.29 & 0.707 & \underline{0.937} & 0.717 & 0.667 & 0.849 & \underline{0.823} & 0.783 \\
         \footnotesize CFM (\citet{tong2023improving})  & 2 & \underline{0.737} & 0.913 & \underline{0.730} & 0.690 &  0.870 & 0.797 & 0.790 \\
        \footnotesize OT-CFM (\citet{tong2023improving}) & 2 & 0.667 & 0.897 & 0.613 & 0.630 & 0.694 & 0.753 & 0.709 \\
        \midrule
        \footnotesize COT Policy (Ours) & 2 & \textbf{0.750} & \underline{0.937} &\textbf{0.787} & \underline{0.707}  &\underline{0.878} & \textbf{0.847} & \textbf{0.818} \\
        \bottomrule
    \end{tabular} 
    }
    \vskip 0.1in
    \caption{Performance comparison of COT Policy with all baselines on the robot manipulation tasks.}
    \label{tab:manipulation_tasks}
\end{table}

\subsection{Simulated Robot Manipulation Tasks}
\label{sec:experiments_robotsim}

We assess the performance of COT Policy in robot manipulation tasks. We choose four tasks from the MimicGen benchmark along with \texttt{push-t} and \texttt{cup}. We evaluate all policies on 150 rollouts and 2 different seeds for the noise distribution, except for \textit{Push-T}, which is evaluated on 600 rollouts in total. For each task we report the average success rate over all rollouts and seeds, except \texttt{push-t}, for which we report the average percentage of coverage of the goal from the T block, consistent with evaluation metrics from prior work. For Adaflow we use the trained CFM policy and fine-tune the variance predictor as explained in \citep{hu2024adaflow}. For CFM, OT-CFM, and COT Policy we use the \texttt{midpoint} solver to sample action trajectories, for Diffusion Policy we use DDIM~\citep{song2020denoising}, and for Adaflow we use the variance adaptive solver with 2 maximum steps to have a fair comparison with COT Policy.

\paragraph{2-step COT Policy outperforms all baselines.} From \Cref{tab:manipulation_tasks} it can be seen that COT Policy outperforms all compared methods on average. COT Policy achieves the highest success rate across tasks with $\text{NFE}=2$ and even exceeds CFM, OT-CFM, and DP when those models are evaluated with $\text{NFE}=4$ or DP with $\text{NFE}=20$. OT-CFM again lags behind all the other methods since it does not take into account the observations when coupling action trajectories with noisy trajectories, yielding a biased optimal flow. We hypothesize that Adaflow performs worse than 2-step CFM, despite having an adaptive step size,  because it takes a maximum of 2 \texttt{euler} steps instead of the 2 steps that are needed for one iteration of the \texttt{midpoint} method in the other flow-based policies. 

\subsection{Evaluating Multimodality}

\label{sec:experiments_multimodality}

We further evaluate the diversity of actions generated by COT Policy and by a CFM policy. To measure trajectory diversity, we introduce the \textit{Trajectory Variance ($TV$)} metric. 
For a set of $n$ trajectories $\mathcal{A} = \{a_1, a_2, \dots, a_n\}$, we calculate $TV$ as the mean squared Dynamic Time Warping \citep[DTW;][]{muller2007dynamic} distance between each trajectory in $\mathcal{A}$ and the barycenter (or Fr\'echet mean) of $\mathcal{A}$, calculated using DWT Barycentric Average \citep[DBA;][]{petitjean2011global}, as implemented in the \textit{tslearn} package~\citep{tavenard2020tslearn}. 
The barycenter $\mu_a$ acts as a `mean trajectory' and is defined as $\mu_a = \arg\min_{\mu} \sum_{i=1}^{n} d_{\text{DWT}}^2(a_i, \mu)$, where $d_{\text{DWT}}$ is the DWT distance between two trajectories. 
Using $\mu_a$, we calculate $TV$ as $TV=\frac{1}{n}\sum_{i=1}^{n}d_{\text{DWT}}^2(a_i, \mu_a)$. We view $TV$ as an analogue of variance in the space of trajectories and use it to measure the diversity of action plans generated by the two policies. (In the case of one-step trajectories with the same initial point, $TV$ exactly recovers the trace of the empirical covariance matrix of the velocity vectors.)

\begin{figure}[t]
    \centering
    \includegraphics[width=\linewidth]{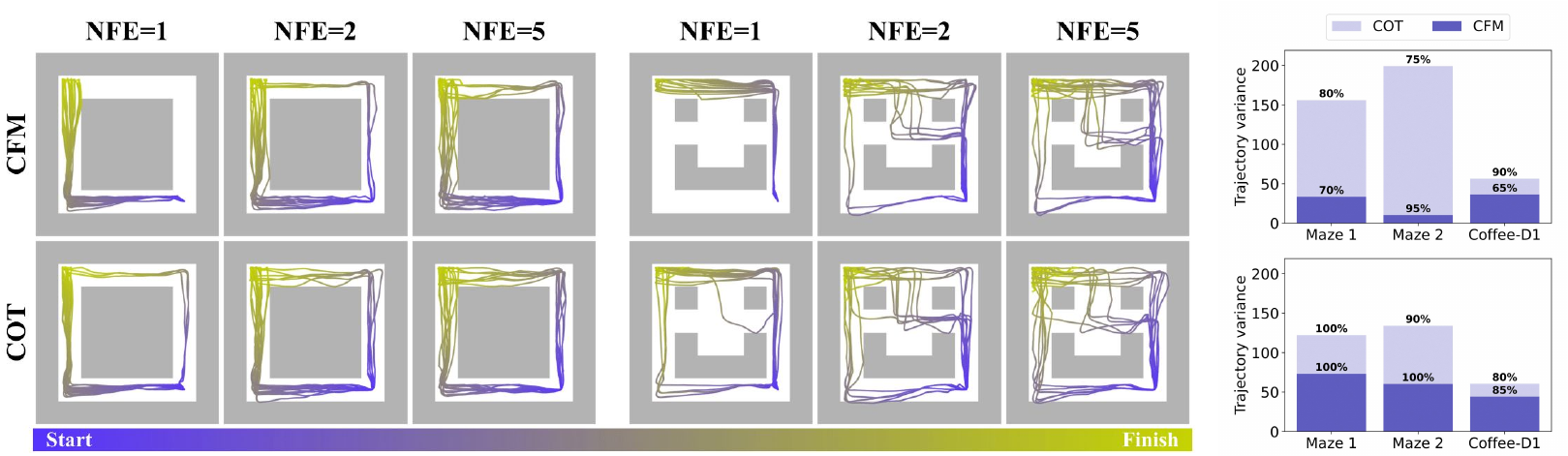}
    \caption{\textbf{Left:} Evaluation of the ability of CFM and COT Policy to encode multiple modes in the maze navigation task for 1, 2, and 5 \texttt{euler} steps (NFE). \textbf{Right:} Trajectory Variance of CFM and COT Policy in the two Maze tasks and \texttt{coffee\_d1} task for $\text{NFE}=1$ on the top and $\text{NFE}=2$ on the bottom. The percentage of successful rollouts is written above each bar.}
    \label{fig:cot_vs_cfm_multimodal}
\end{figure}

Out of 50 rollouts with the same environment seed, only successful ones are considered for the calculation of $TV$ in the results presented in \Cref{fig:cot_vs_cfm_multimodal}.

\paragraph{COT Policy generates diverse actions even for low NFE.} Based on the results shown in \Cref{fig:cot_vs_cfm_multimodal}, COT Policy exhibits more diversity (higher $TV$) for all tasks, which is in accordance with the trajectories shown on the left side of the same figure. The $TV$ and success rate of CFM increases when we use 2 \texttt{euler} steps instead of 1, while for COT Policy we observe an average increase in success rate but slightly lower $TV$. This is potentially due to discarding unsuccessful rollouts in the calculation of $TV$. Discarding certain trajectories can lead to either an increase or decrease of $TV$ depending on which mode they represent.
See \Cref{sec:tv_results} for additional results.

\subsection{Ablations}

\begin{wrapfigure}{r}{0.35\textwidth}
    \vspace*{-4em}
    \centering
    \includegraphics[width=\linewidth]{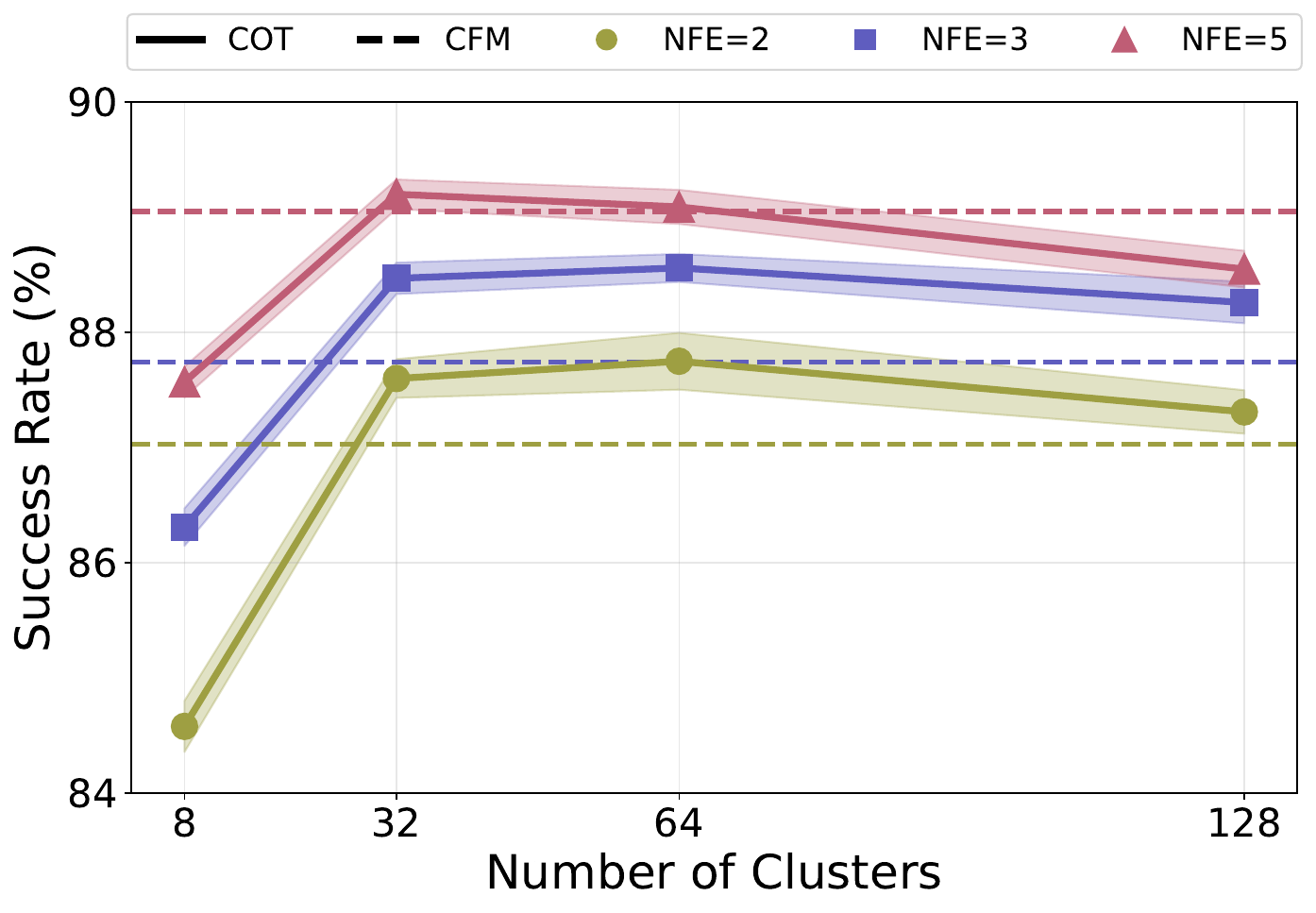}
    \caption{Success rates of COT Policy on the \texttt{push-t} task for $K=8, 32, 64, 128$ and $\text{NFE}=2,4,10$.}
    \label{fig:performance_v_num_clusters}
    \vskip -2em
\end{wrapfigure}
\paragraph{Effect of cluster number in COT Policy.} The number of clusters $K$ used in COT Policy directly impacts the performance. From \Cref{fig:performance_v_num_clusters} it is evident that COT with too few clusters exhibits similar degradation to OT-CFM. For a number of clusters close to the batch size we observe the greatest advantage of COT over CFM. 
This is potentially due to the diversity of clusters in each batch this choice induces, which should not be too low (exhibiting close to OT-CFM behavior) or too close to the batch size (exhibiting close to CFM behavior as each condition is unique). As the number of clusters increases and approaches the dataset size, the behavior of COT-CFM asymptotically resembles that of CFM. Further discussion can be found in \Cref{sec:number_of_clusters}. Additionally, we show the importance of observation discretization by providing a comparison of COT Policy with and without clustering in \Cref{sec:appendix_cot_wo_clustering}.

\begin{figure}[t]
    \includegraphics[width=\textwidth]{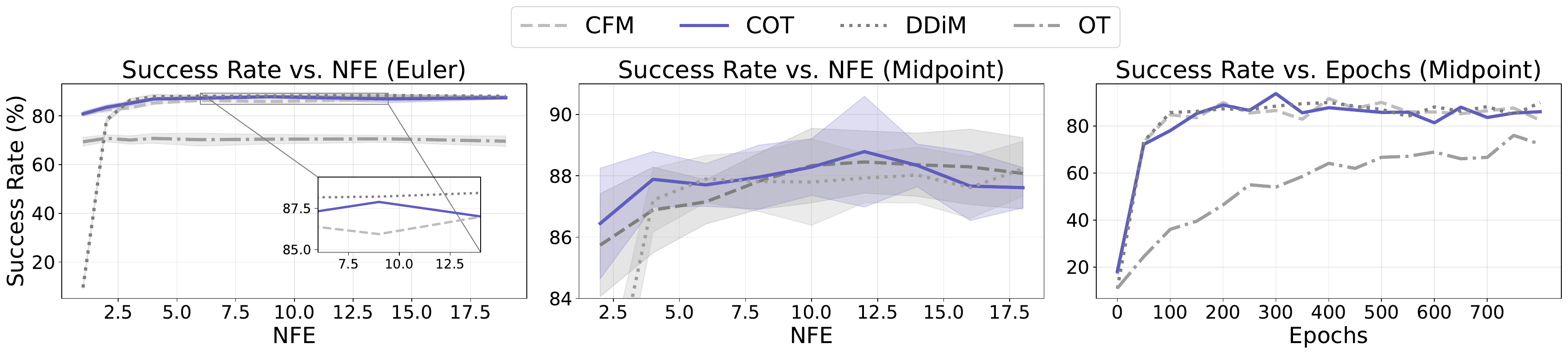}
    \caption{Success rates on the \texttt{push-t} task with varying NFE with the \textbf{Left:} \texttt{euler} solver and \textbf{Middle:} \texttt{midpoint} solver. \textbf{Right:} success rate over training epochs.}
    \label{fig:ablation}
\end{figure}

\paragraph{Effect of NFE and solver type.} From \Cref{fig:ablation} and the rest of the results presented in this section, it is clear that NFE has a direct impact on the quality and diversity of the generated samples from any method. 
The number of solver steps, despite heavily impacting all methods, has a more subtle effect on COT Policy as sample diversity and success rates only slightly benefit from increases in NFE. 
Furthermore, the choice of the fixed-step solver used plays a role according to \Cref{fig:ablation} as there is a consistent increase in performance with the \texttt{midpoint} solver for the same NFE.

\paragraph{Training complexity.} In \Cref{fig:ablation}, we see that COT Policy training converges at approximately the same time as CFM and DP. 
Moreover, our choices for the encoder $\mathcal{E}$ and the discretizer $\mathcal{Q}$ (PCA and K-means respectively) as well as minibatch OT add negligible computational overhead per batch when implemented in GPU-accelerated frameworks, compared to performing a backward pass on the flow model. 
Therefore, COT Policy requires approximately the same time as CFM and DP to train.

\subsection{Real Robot Experiments}
\label{sec:real_robots}

\begin{wrapfigure}{r}{0.42\textwidth}
    \vspace*{-4em}
    \begin{minipage}{\linewidth}
        \centering
        \includegraphics[width=\linewidth]{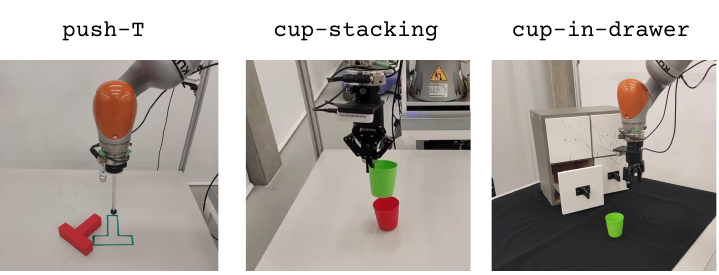}
        \captionof{figure}{Visualization of real tasks.}
        \label{fig:real_tasks}
        \vspace{0.3em}
        \small
        \resizebox{\linewidth}{!}{
        \begin{tabular}{@{}l cc cc}
        \toprule
        \multirow{2}{*}{Task} & \multicolumn{2}{c}{CFM policy} & \multicolumn{2}{c}{COT policy} \\
        \cmidrule(lr){2-3} \cmidrule(lr){4-5}
                              & $SR \uparrow$ & $TTC \downarrow$ &  $SR \uparrow$ & $TTC \downarrow$ \\
        \midrule
        \texttt{push-T}         & $0.2\ /\ 0.4$ & $54.4\ /\ 44.7$ & $\mathbf{0.8}\ /\ 0.6$ & $\mathbf{35.7}\ /\ 35.8$ \\
        \texttt{cup-stacking}   & $0.4\ /\ 0.2$ & $44.5\ /\ 51.1$ & $0.6\ /\ \mathbf{0.8}$ & $32.2\ /\ \mathbf{24.0}$ \\
        \texttt{cup-in-drawer}  & $0.2\ /\ 0.2$ & $53.3\ /\ 54.0$ & $\mathbf{0.8}\ /\ 0.2$ & $\mathbf{35.3}\ /\ 52.9$ \\
        \bottomrule
        \end{tabular}
        }
        \captionof{table}{Success Rate (SR; 5 rollouts) and Time to Completion (TTC; seconds) in the real tasks. Results in the format (\texttt{euler}--NFE=1 / \texttt{midpoint}--NFE=2).}
        \label{tab:real_results}
    \end{minipage}
    \vskip -1em
\end{wrapfigure}

We evaluated the low-NFE performance of COT Policy and CFM Policy on three real-world robot tasks, shown in \Cref{fig:real_tasks}, using a KUKA IIWA 14 robot and two Realsense D415 cameras---one mounted on the end-effector and one providing external view of the workspace.
We collected expert demonstrations using teleoperation and executed the policy in a PC with a NVIDIA GeForce RTX 2080.

In all three tasks, CFM underperforms compared to COT Policy, while also requiring more time to solve the task (\Cref{tab:real_results}). 
The increased NFE required by CFM to accurately generate actions from the multimodal distribution of expert human demonstrations leads to intermittent motions.
In contrast, COT policy is able to solve all tasks with at most 2 inference steps. 

\section{Conclusions}
We present COT Policy, a flow matching policy for fast inference in robot tasks and multimodal action generation. 
We achieve this by pairing noise and action trajectories during training with an approximation to the Conditional Optimal Transport plan. 
We have shown that 2-step COT Policy outperforms other generative models in sequential decision making tasks while maintaining action diversity.
This allows COT Policy to be used for high-quality real-time action generation while requiring only a handful of network evaluations to perform inference.

\clearpage
\section{Limitations} One major limitation of our method is the introduction of two hyperparameters. 
Although the scale factor $\gamma$ does not significantly impact performance, the choice of cluster $K$ does. 
Different dataset and batch sizes can impact the optimal choice of $K$. 
For robot manipulation tasks we suggest setting $K$ equal to the batch size however it is not evident if that choice is sensible for any task.
Additionally, although we are able to perform high quality 2-step action generation with the \texttt{midpoint} solver, in some cases there is a gap between low and high NFE performance, especially as dimensionality grows. 
It would be an interesting direction to address this gap by combining OT coupling with distillation techniques that require a single training phase~\citep{frans2024one, yang2024consistency}. 

\bibliography{main}

\clearpage
\appendix

\section{Related Work}
\label{sec:rw} 

\paragraph{Diffusion and flow-based models in robotics.}  
Diffusion and flow-based models have demonstrated remarkable performance across various domains, ranging from image and video generation~\citep{rombach2022high, saharia2022photorealistic, jin2024pyramidal, blattmann2023align} to monocular depth estimation~\citep{ke2024repurposing, saxena2024surprising, gui2024depthfm}.  
Their ability to model complex, high-dimensional probability distributions has naturally led to their widespread use in robotics, particularly in policy learning~\citep{urain2024deep, ajay2022conditional}.  
\citet{chi2023diffusion} proposed DMs as an alternative to standard explicit policy models~\citep{mandlekar2021matters}, learning distributions of short-horizon trajectories conditioned on the robot's proprioceptive and exteroceptive observations.  
Other methods~\citep{carvalho2023motion, janner2022planning} have leveraged DMs to develop accurate planners that integrate rewards into trajectory generation through conditional generation techniques, such as diffusion guidance~\citep{ho2022classifier}.  

Recent work has explored the effectiveness of generative model (GM)-based policies conditioned on robot-relevant observations, including point clouds \citep{chisari2024learning}, text \citep{xian2023chaineddiffuser, zhang2024affordance}, and force-torque data \citep{hou2024adaptive, aburub2024learning}.  
In many cases, the geometric structure of the observations has been explicitly considered by incorporating manifold variants of Flow Matching \citep{ding2024fast, chisari2024learning} or employing equivariant Transformer architectures for policy learning \citep{funk2024actionflow}.  

Despite their success in various robotic tasks, DMs and flow-based models suffer from slow inference speeds, posing a significant limitation for real-time deployment or scenarios with constrained computational resources.  
Some approaches leverage FM to reduce the number of steps required for action sampling. However, even in flow-based policies, few-step generation can significantly impact the diversity of action trajectories and, potentially, the policy's success rate.  

\paragraph{Accelerated inference in continuous-time generative models.}  
Efforts to accelerate inference in DMs began with the development of alternative samplers \citep{song2020denoising, lu2022dpm, karras2022elucidating}, which replace computationally expensive iterative denoising with more efficient, often deterministic, sampling schemes.  
While these methods significantly improve inference speed, they still require multiple steps to generate high-quality samples.  
To achieve competitive one-step inference quality, distillation methods \citep{salimans2022progressive, yang2024consistency, song2023consistency, kim2023consistency, frans2024one, yin2024one, liu2022flow} are commonly employed.  

This family of approaches primarily seeks models capable of directly predicting the solution to the denoising SDE at any time step, enforcing consistency objectives \citep{song2023consistency, kim2023consistency, frans2024one} rather than explicitly distilling few-step models using generated denoising SDE solutions.  
A major drawback of distillation methods is their reliance on costly training procedures, often requiring multiple training phases involving teacher DMs and student one-step models.  
Some high-quality one-step models \citep{salimans2022progressive, liu2022flow} necessitate progressive training phases.  
While certain methods can be trained in a single phase \citep{song2023consistency, frans2024one}, they require more training steps to converge, each step being computationally expensive, and the performance gap between few-step and one-step inference remains large.  
Most robot policies based on distillation methods \citep{prasad2024consistency, wang2024one, lu2024manicm} face similar limitations.   

In this work, we build upon Conditional Optimal Transport principles to develop policies that train as efficiently as CFM while achieving superior few-step performance.  \textit{Our method serves as an efficient alternative to CFM by implicitly improving few-step performance via learning straighter paths. This is in contrast to distillation methods which explicitly learn few-step solutions of the denoising ODE.}

\section{Implementation Details: Simulation}
\label{sec:implementation}

\subsection{Training details}
\label{sec:appendix-training-details}
\paragraph{Training on manipulation and Maze tasks.} All the baselines have been trained using the same CNN-based U-Net architecture from~\citep{chi2023diffusion}. The flow network for all flow-based techniques and the noise network of DP utilize the U-Net as their architecture, with the same hyperparameters, totaling approximately 240 million parameters (excluding the vision encoders). The image observations first pass through a vision encoder. For more details on the vision encoder see \Cref{sec:appendix-visual-encoder}. We detail further training decisions below:
\begin{itemize}[left=0pt,nosep]
    \item \textbf{CFM}. For the implementation of the CFM-based policy we followed the methodology of~\citet{tong2023improving}. The only discrepancy is the addition of the observations as a conditioning variable in the flow model. 
    \item \textbf{OT-CFM}. OT-CFM uses the same setup as CFM with the addition of coupling noise and action samples using OT. For that we use the tools provided by \textit{torchcfm}~\citep{tong2023improving, tong2023simulation} for calculating the minibatch OT plan with conditions. Since the noise and action samples are rearranged based on the OT plan, the conditions are rearranged as well with the respective actions. To calculate the OT plan we use the Earth Mover's Distance (EMD) algorithm~\citep{flamary2021pot}.
    \item \textbf{Adaflow}. We use the trained CFM models as the base flow models of Adaflow for each task. We then train the variance estimation network, implemented as a 2 layer MLP, according to the training procedure detailed in~\citep{hu2024adaflow}, for an additional 200 epochs. The flow network was frozen during training of the variance estimation network. Since we were unable to find a recommended number of training epochs for the variance estimation network we opted for a large enough epoch count despite updating only the weights of some linear layers. The learning rate was held constant throughout training. We set $\eta=0.5$ and $\epsilon_{min}=2$. The choice of $\epsilon_{min}=2$ was made so that Adaflow uses a maximum of 2 steps during inference, to ensure a fair comparison with other 2-step methods. This value of $\eta$ was chosen since slightly higher values led to Adaflow using NFE=2 and slightly lower NFE=1 at every inference step. 
    \item \textbf{DP}. The implementation of DP is identical to the one in~\citet{chi2023diffusion}.
\end{itemize}

\begin{table}[h]
    \centering
    \resizebox{\linewidth}{!}{
    \begin{tabular}{@{}l|cc|cc|cccc}
        \toprule
        Hyperparameter & \multicolumn{2}{c}{Moons \& Fork} & \multicolumn{2}{c}{Maze} & \multicolumn{3}{c}{MimicGen, Push-T \& Ball-in-cup} \\
        \cmidrule(lr){2-3} \cmidrule(lr){4-5} \cmidrule(lr){6-8}
         & CFM \& OT-CFM &  COT Policy & CFM \& OT-CFM &  COT Policy  & DP & CFM \& OT-CFM &  COT Policy  \\
        \midrule
        Learning rate & 1e-3 & 1e-3 & 1e-4 & 1e-4 & 1e-4 & 1e-4 & 1e-4 \\
        Optimizer & Adam & Adam & AdamW & AdamW & AdamW & AdamW & AdamW \\
        Batch size & 256 & 256 & 64 & 64 & 64 & 64 & 64 \\
        Weight decay & 0.0 & 0.0 & 1e-6 & 1e-6 & 1e-6 & 1e-6 & 1e-6 \\
        Epochs & -- & -- & 1000 & 1000 & 1000 & 1000 & 1000\\
        Training Steps & 50000 & 50000 & -- & -- & -- & -- & -- \\
        Learning rate schedule & constant & constant & cosine & cosine & cosine & cosine & cosine \\
        EMA decay rate & -- & -- & 0.9999 & 0.9999 & 0.9999 & 0.9999 & 0.9999 \\
        Warmup steps & -- & -- & 1500 & 1500 & 1500 & 1500 & 1500 \\
        \midrule
        Inference steps in training rollouts & - & - & 2 (Euler) & 2 (Euler) & 100 (DDIM) & 2 (Midpoint) & 2 (Midpoint) \\
        Action prediction horizon & - & - & 60 & 60 & 16 & 16 & 16 \\
        Observation horizon & - & - & 2 & 2 & 2 & 2 & 2 \\
        Action execution horizon & - & - & 50 & 50 & 8 & 8 & 8 \\
        Image size & 2(Moons)/1(Fork) & 2(Moons)/1(Fork) & 96x96 & 96x96 & 96x96 & 96x96 & 96x96 \\
        \midrule
        Number of Clusters & -- & 2 & -- & 16 & -- & -- & 64 \\
        PCA features & -- & -- & -- & 100 & -- & -- & 100 \\
        \bottomrule
    \end{tabular}
    }
    \caption{Training hyperparameters for COT Policy and baselines}
    \label{tab:training_hyperparameters}
\end{table}
\paragraph{Training on distribution tasks.} We train CFM, OT-CFM, and COT-CFM using an MLP with 4 Linear layers with 64 neurons each and SELU activations. The conditioning and time variables are concatenated with the samples and passed as inputs. For these two tasks we do not use positional embeddings for the time variable. For the Fork task we do not encode the conditions as they are already 1-dimensional, however we cluster them using K-Means with $K=2$. We found that this number of clusters works best for this task, potentially due to the fact that the distribution changes based on two different sets of values for the condition variable, namely $x\leq0$ and $x>0$. 

\subsection{Clustering and dimensionality reduction in COT Policy}
\label{sec:appendix-clustering}
As mentioned in the main text, for COT Policy, we implement the encoder $\mathcal{E}$ as a PCA transformation of the images and the discretizer $\mathcal{Q}$ as K-means clustering, with the centroids of the clusters being the discretized values of the conditions. Both algorithms were implemented in \textit{pytorch} to allow GPU acceleration. The eigenvector matrix for PCA is calculated using SVD on the images from the entire dataset before training and during training dimensionality reduction is performed by multiplying the data matrix with the eigenvector matrix. Similarly the clusters are computed using the entire dataset before training and during training each pair of PCA image features and proprioception is matched to its closest cluster, based on Euclidean distance. Although the speed of these operations is negligible compared to forward and backward passes of the U-Net, the storage of the eigenvector matrix for PCA can have a significant impact on the GPU memory required during training. 

\subsection{Visual Encoder}
\label{sec:appendix-visual-encoder}
We train the visual encoder used for generating the embeddings that condition the flow, end-to-end alongside the flow network, using a randomly initialized ResNet-18~\cite{he2016deep}.  
This choice aligns with~\cite{chi2023diffusion}, which reported that conditioning a DP with features extracted from pre-trained visual representations (PVRs) (\eg, R3M~\cite{nair2022r3m}) led to suboptimal performance.  
However, recent advances in 3D spatial representations, fused with PVR-derived features while ensuring multi-view consistency, have demonstrated promising results in downstream robotic tasks~\cite{tsagkas2023vlfields, tsagkas2024click, li2024affgrasp, huang2023voxposer, clip-fields}.  
This has sparked interest in leveraging such methods for conditioning diffusion policies~\cite{3d_diffuser_actor, xian2023chaineddiffuser}.  
We argue that this research direction is still in its early stages, with open challenges in effectively integrating PVR features into imitation learning policies~\cite{tsagkas2025pretrainedvisualrepresentationsfall}.  
In future work, we aim to investigate how PVRs can enhance policy generalization, particularly in out-of-distribution scenarios.

\subsection{Demonstration collection}
\label{sec:demonstration_collection}

\paragraph{MimicGen.} For all MimicGen tasks we use the datasets provided by the authors of \citep{mandlekar2023mimicgen}. These datasets consist of 1000 demonstrations, synthetically generated from a total of 10 human demonstrations. We only keep the first 100 demonstrations to keep training times within reasonable limits. Synthetically generated datasets can have an immediate impact on the NFE needed to get the optimal performance, as demonstrations collected using deterministic policies lack multimodality. Lack of multimodality implies conditional distributions that lack variance and therefore few-step CFM and COT-CFM performance should be similar. However, we found that for the MimicGen tasks considered in this paper there was enough variation in the demonstrations. Ideally, human-collected demonstrations would have the maximum variation, however we avoided using the tasks and demonstrations used in~\citet{chi2023diffusion}, as most generative policies solve the majority of them with $100\%$ success rate. 

\paragraph{PushT and BallInCup.} For the \texttt{push-t} task we used 90 demonstrations provided in \citet{chi2023diffusion}. For the \texttt{cup} task, we trained a PPO~\citep{schulman2017proximal} agent and collected 100 demonstrations by executing the learned policy. 

\paragraph{Maze2D.} We used the policy that comes with D4RL~\citep{fu2020d4rl} for collecting demonstrations in the Maze environments, implemented as a q-iteration policy combined with a waypoint controller. We modified the policy by adding noise in the $q$ values in order to generate diverse demonstrations when the starting and goal positions are fixed.  A total of 100 demonstrations were collected for each maze environment.

\subsection{Hardware}
\label{sec:hardware}

Training and evaluation for the simulated tasks were performed using a workstation with (CPU, RAM, GPU): AMD Ryzen Threadripper PRO 5965WX 24-Cores, 128GB, $2\times$ NVIDIA GeForce RTX 4090.

\section{Implementation Details: Real-robot tasks}
\label{sec:implementation_real}

\subsection{Training details}
\label{sec:appendix_training_details_real}
For all real-robot task and for both CFM and COT policies we use the U-Net architecture for the flow as described in \Cref{sec:appendix-training-details}. The visual encoders were also trained end-to-end and followed the same architecture described in \Cref{sec:appendix-visual-encoder}, with the only difference being the resolution of the images used. We decided to use higher resolution images for \texttt{push-T} (\eg, 240x320 opposed to 96x96 (or 96x120) used in simulation and real-robot manipulation tasks) in order to accurately capture the pose of the T block and the target. For more details on the hyperparameters used for training on the real-robot tasks, see \Cref{tab:training_hyperparameters_real}.

\begin{table}[h]
    \centering
    \resizebox{\linewidth}{!}{
    \begin{tabular}{@{}l|cc}
        \toprule
        Hyperparameter & \multicolumn{2}{c}{\texttt{push-T}, \texttt{cup-stacking}, \texttt{cup-in-drawer}} \\
        \cmidrule(lr){2-3}
         & CFM Policy & COT Policy \\
        \midrule
        Learning rate & 1e-4 & 1e-4 \\
        Optimizer & AdamW & AdamW \\
        Batch size & 64 & 64 \\
        Weight decay & 1e-6 & 1e-6 \\
        Epochs & 1000 & 1000 \\
        Learning rate schedule & cosine & cosine \\
        EMA decay rate & 0.9999 & 0.9999 \\
        Warmup steps & 1500 & 1500 \\
        \midrule
        Action prediction horizon & 16 & 16 \\
        Observation horizon & 2 & 2 \\
        Action execution horizon & 8 & 8 \\
        Image size &  240x320 (\texttt{push-T}) / 96x120 (others) & 240x320 (\texttt{push-T}) / 96x120 (others)\\
        \midrule
        Number of Clusters & -- & 64 \\
        PCA features & -- & 100 \\
        \bottomrule
    \end{tabular}
    }
    \caption{Training hyperparameters for CFM Policy and COT Policy across real-robot tasks.}
    \label{tab:training_hyperparameters_real}
\end{table}

\subsection{Demonstration collection}
\label{sec:appendix_demonstration_collection_real}
For each of the three tasks CFM and COT policies were tested on using real hardware, we collected a total of 30 demonstrations using a 3D spacemouse. The actions used for learning the two policies were end-effector twists that were commanded through the teleoperation device. For the \texttt{push-T} task we only allowed translational motion of the end-effector, making the actions 2-dimensional. For the rest of the tasks the actions were 7-dimensional (6 dimensions for controlling the twist of the end-effector and 1 for controlling the gripper). 

Despite having less demonstrations for the real tasks than for the simulation tasks, the real datasets ended up highly multimodal and covered multiple initial configurations of the workspace and the robot. 

\subsection{Evaluation details}
\label{sec:appendix_evaluation_details}

\paragraph{Test protocol}We evaluated the performance of both CFM and COT policies on 5 random environment configurations for each task. The environment configurations were chosen a-priori and were recreated as accurately as possible after every rollout. We further made sure that none of these environment configurations appeared identical in our datasets. 

For each task we tested both policies using the minimum amount of steps for each of the \texttt{euler} and \texttt{midpoint} solvers (\eg, 1 and 2 respectively). The maximum allowed time before considering a rollout a failure was chosen to be 1 minute. For each of the unsuccessful rollouts reported in \Cref{tab:real_results}. their contribution to the average $TTC$ was 60 seconds. 

\begin{figure}[h!]
\begin{center}
\centerline{\includegraphics[width=\columnwidth]{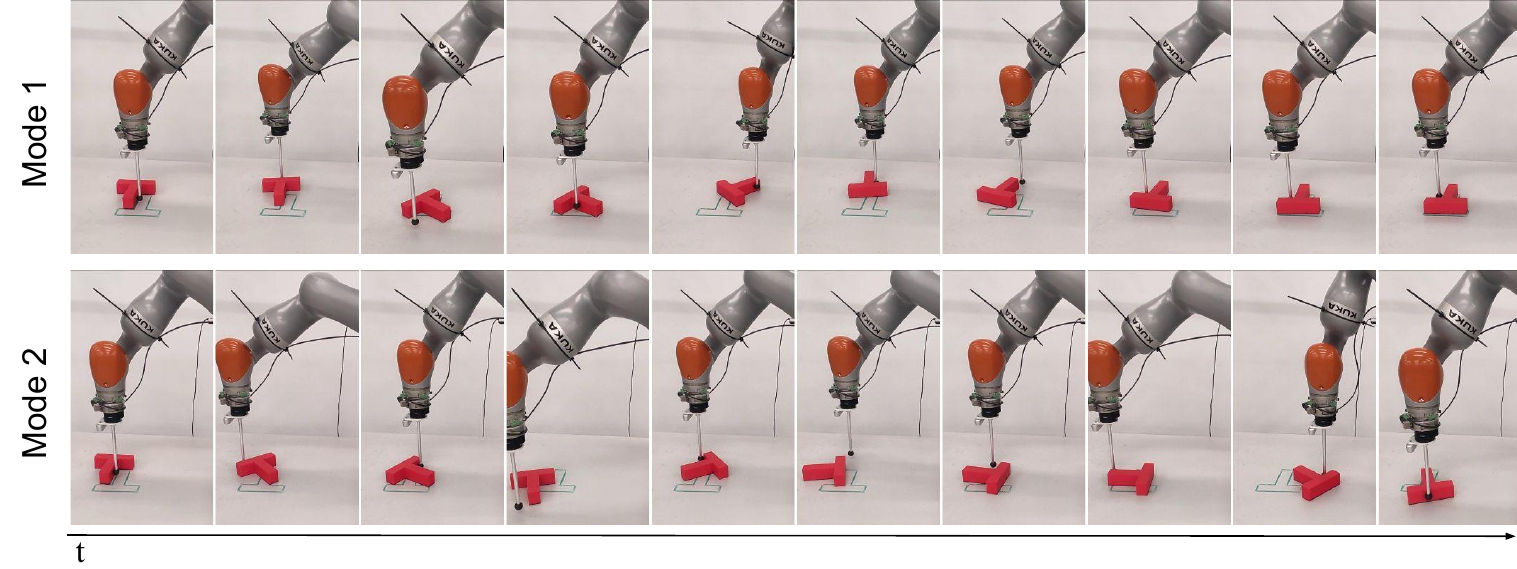}}
\caption{Two modes of solving the real \texttt{push-T} task (\eg, by rotating the T clockwise and anti-clockwise) uncovered by COT Policy with NFE=1 using the \texttt{euler} solver. } 
\label{fig:modes_pusht}
\end{center}
\end{figure}

\paragraph{Discussion}Throughout our experiments we have consistently found that COT policies outperformed CFM policies for low-NFE inference. Moreover, we empirically observed that CFM policies would often get stuck for extended periods of time or produce jerky motions. This was less frequent when the \texttt{midpoint} solver was used, however success rates were still low. On the contrary, COT Policy succeeded for most of the initial configurations in all tasks. We have observed that COT policy was able to reproduce most of the modes existing in the dataset, even with 1-step \texttt{euler} inference, as can be seen in \Cref{fig:modes_pusht}, where the robot pushes the T block in the target with both clockwise and anti-clockwise motions. 

Although performance is similar when using the \texttt{euler} and the \texttt{midpoint} solver, there is a big gap in the \texttt{cup-in-drawer} task. We suspect that this gap mainly stems from the intermittent motions induced by the 2-step \texttt{midpoint} on a task that is dynamic and has a high risk of collision. Low performance for CFM policy is consistent with the other tasks and stems from the inability of CFM to generate high quality samples with few-step inference and the reasons described above when it comes to inference with the \texttt{midpoint} solver. Intuitively, intermittent motion for few-step inference could be mitigated by having a faster GPU on the robot's control computer, however this is not always possible.

\subsection{Hardware}
\label{sec:appendix_hardware_real}

Training for the real tasks was performed using a workstation with (CPU, RAM, GPU): AMD Ryzen Threadripper PRO 5965WX 24-Cores, 128GB, $2\times$ NVIDIA GeForce RTX 4090. Inference on the robot was performed using a desktop PC with (CPU, RAM, GPU): Intel(R) i9-9900KF, 64GB, NVIDIA GeForce RTX 2080.

\section{Additional evaluation results}
\subsection{Low-Dimensional Distribution Learning}
\label{sec:experiments_synthetic}

\begin{wraptable}{r}{0.48\columnwidth}
    \centering
    \small
    \vspace{-30pt}
    \resizebox{\linewidth}{!}{
    \begin{tabular}{@{}l l cccccc}
        \toprule
        \multirow{2}{*}{Task} & \multirow{2}{*}{} & \multicolumn{2}{c}{CFM} & \multicolumn{2}{c}{OT-CFM} & \multicolumn{2}{c}{COT-CFM} \\
        \cmidrule(lr){3-4} \cmidrule(lr){5-6} \cmidrule(lr){7-8}
        & & $W_2^2 \downarrow$ & NFE & $W_2^2 \downarrow$ & NFE & $W_2^2 \downarrow$ & NFE \\
        \midrule
        \multirow{3}{*}{\shortstack[l]{Moons \\ 2D}} & \multirow{2}{*}{\textit{E}} & 1.490 & 1 & 0.777 & 1 & \textbf{0.345} & 1 \\
         & & 1.094 & 2 & 0.709 & 2 & \textbf{0.322} & 2 \\
        \cmidrule(lr){2-2}
        & \textit{D} & 0.335 & 134 & 0.683 & 74 & \textbf{0.253} & \textbf{62} \\
        \midrule
        \multirow{3}{*}{\shortstack[l]{Fork \\ 1D}} & \multirow{2}{*}{\textit{E}} & 1.134 & 1 & 2.678 & 1 & \textbf{0.349} & 1 \\
        & & 0.575 & 2 & 0.992 & 2 & \textbf{0.125} & 2 \\
        \cmidrule(lr){2-2}
        & \textit{D} & 0.157 & 74 & 0.981 & \textbf{62} & \textbf{0.072} & 74 \\
        \bottomrule
    \end{tabular}}
    \caption{2-Wasserstein distance on the moons and fork distributions evaluated using \texttt{euler} (\textit{E}) and \texttt{dormand-prince} (\textit{D}) solvers.}
    \vspace{-12pt}
    \label{tab:distribution_tasks}
\end{wraptable}

\begin{figure}[h!]
\begin{center}
\centerline{\includegraphics[width=\columnwidth]{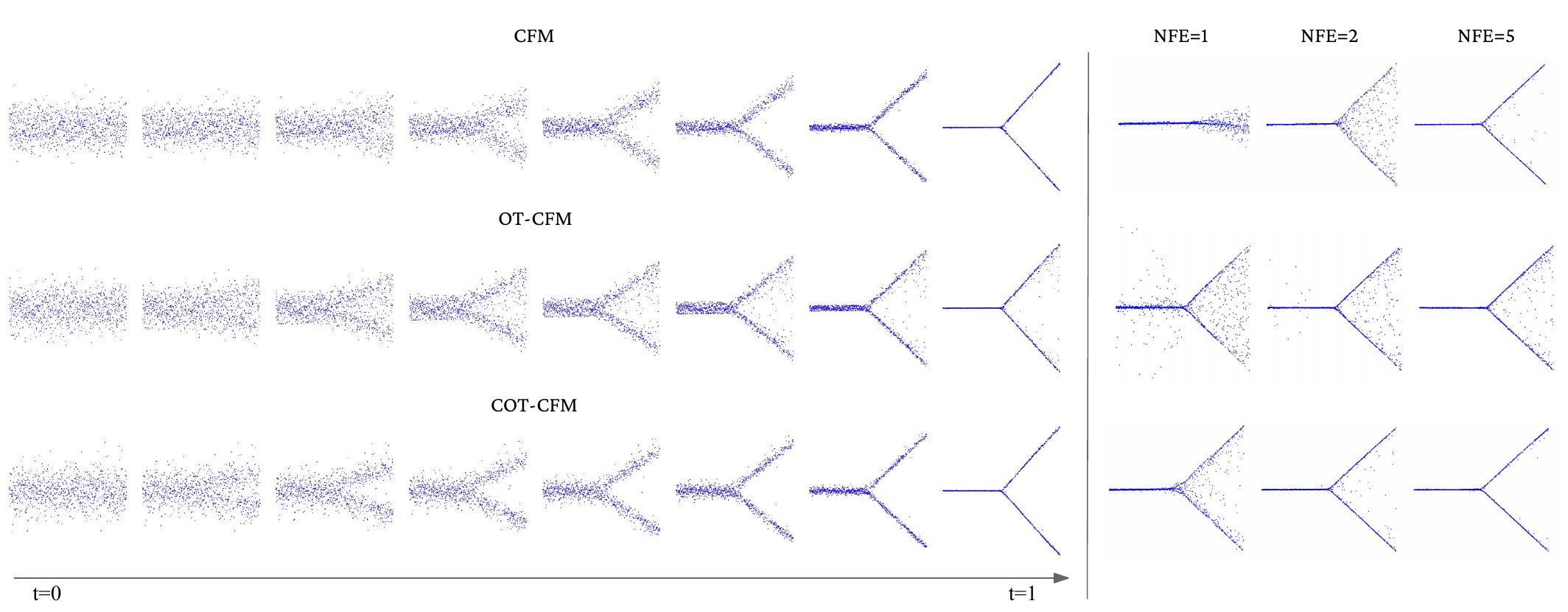}}
\caption{Qualitative performance of CFM, OT-CFM, and COT-CFM on generating the conditional Fork distribution from standard Gaussian noise. \textbf{Left:} The distribution as time progresses from 0 to 1 and \textbf{Right:} the generated distributions for different NFE using the \texttt{euler} solver.}
\label{fig:forks}
\end{center}
\end{figure}

We evaluate how well CFM, OT-CFM, and COT-CFM can recreate the 2D two-moon distribution shown in \Cref{fig:moons} starting from a ring of 8 Gaussian distributions and the 1D fork distribution adapted from \citep{hu2024adaflow}, with a standard Gaussian noise distribution (see \Cref{fig:forks}).
Both distributions are conditional. In the moons distribution, $0$ and $1$ are condition values indicating the top or bottom moon, respectively. 
The fork distribution is a continuous distribution over $y$ conditioned on $x$, defined as $y=0 \text{ if } x\leq 0$ and $y = \mathcal{U}(\{x,-x\}) \text{ if } x>0$. 
Since the condition variable is already low-dimensional, we cluster it into 2 clusters. 

\paragraph{COT-CFM improves sample quality with fewer steps.} We report the 2-Wasserstein distance between the recreated and target distribution in \Cref{tab:distribution_tasks}.  For the same number of few \texttt{euler} steps, COT-CFM outperforms the other two methods. In both tasks, 2-step generation from COT-CFM with an \texttt{euler} solver outperforms $>60$-step CFM and OT-CFM generation with the Dormand-Prince 5th-order (\texttt{dopri5}) solver, an adaptive solver commonly used for solving Neural ODEs. Furthermore, COT-CFM is the only method capable of successfully recreating the two moons for $\text{NFE}=1$ (see \Cref{fig:moons}). This is mainly due to the straight paths the COT couplings induce, which is also evident by the reduced number of steps taken by the \texttt{dopri5} solver. Interestingly, CFM evaluated with a 1-step \texttt{euler} solver approximately generates the conditional mean of the two conditional distributions, while OT-CFM fails to model them accurately even with 100 steps.
These results confirm the claims made in \Cref{sec:conditional_ot_couplings} that (I-)CFM gives unbiased conditional samples with curved integration paths, OT-CFM gives straight paths but biased samples, while the proposed COT-CFM achieves a balance between the two. See \Cref{fig:forks} for further visualizations.

\subsection{Additional Trajectory Variance results}
\label{sec:tv_results}

\begin{figure}[htbp]
    \centering

    \begin{subfigure}{0.32\textwidth}
        \includegraphics[width=\linewidth]{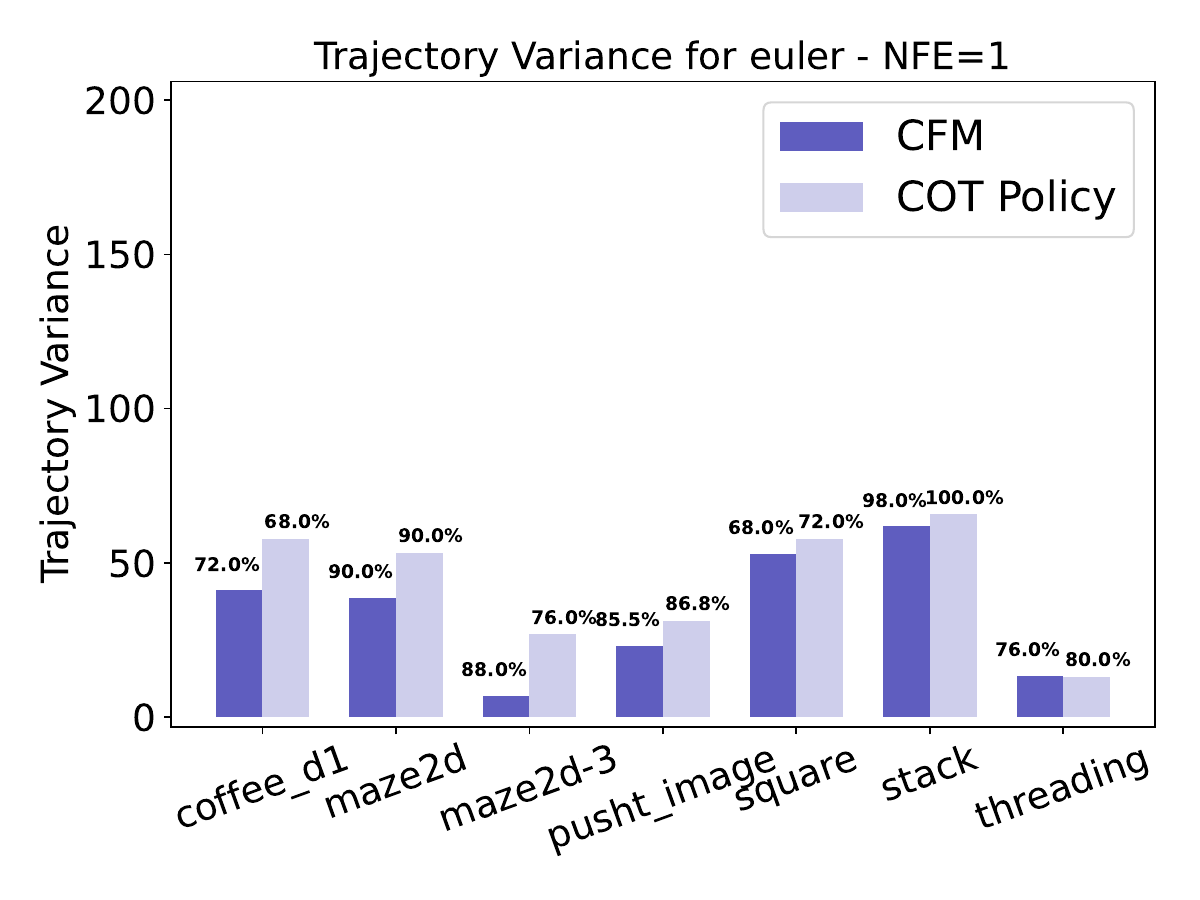}
    \end{subfigure}
    \hfill
    \begin{subfigure}{0.32\textwidth}
        \includegraphics[width=\linewidth]{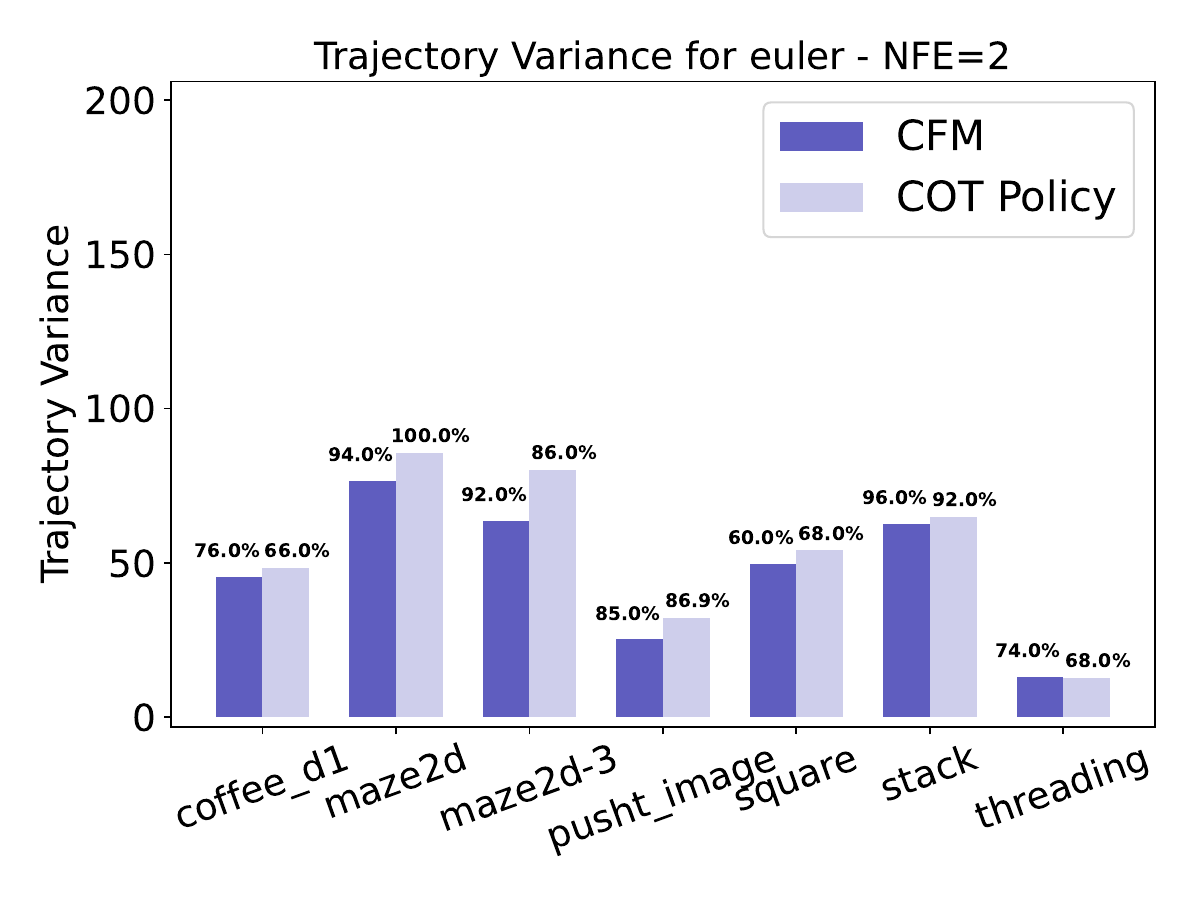}
    \end{subfigure}
    \hfill
    \begin{subfigure}{0.32\textwidth}
        \includegraphics[width=\linewidth]{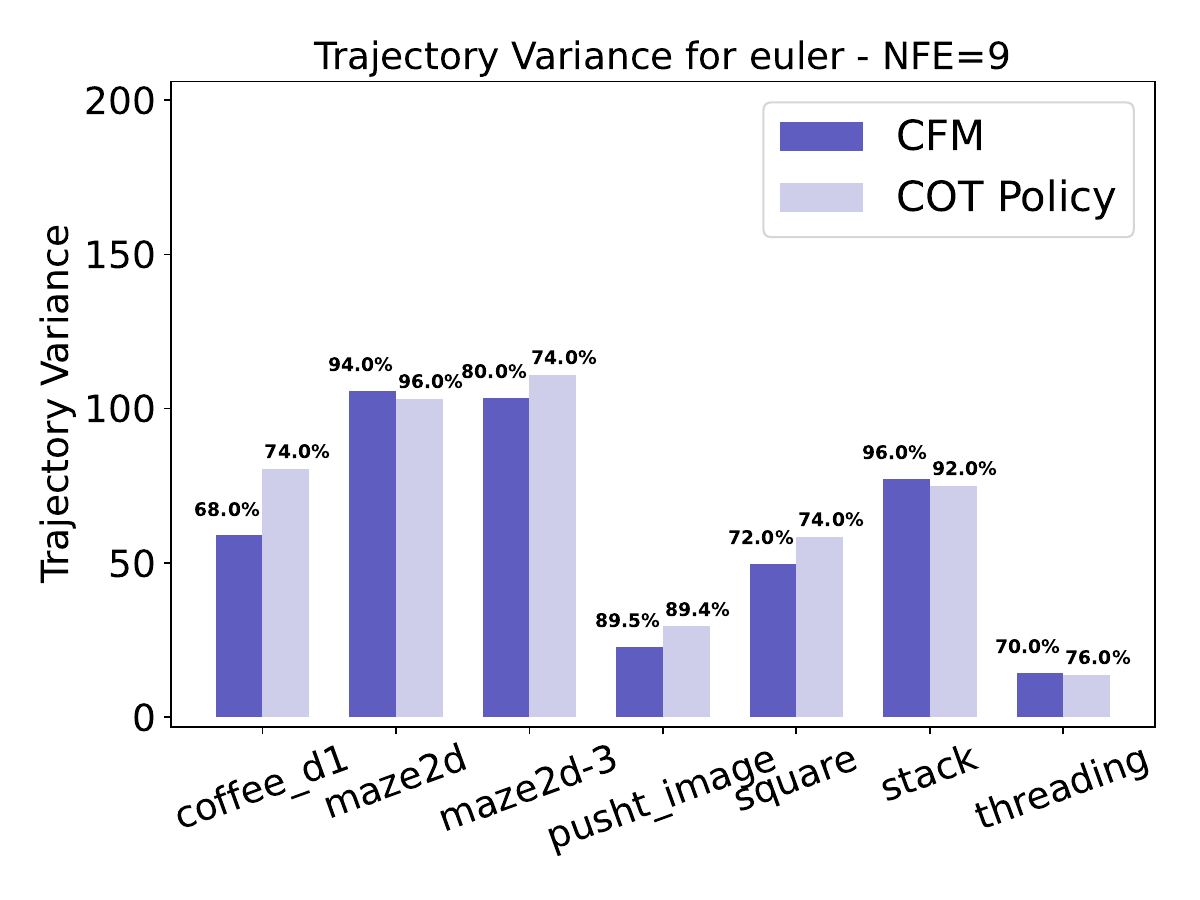}
    \end{subfigure}

    \vspace{0.5cm}

    \begin{subfigure}{0.32\textwidth}
        \includegraphics[width=\linewidth]{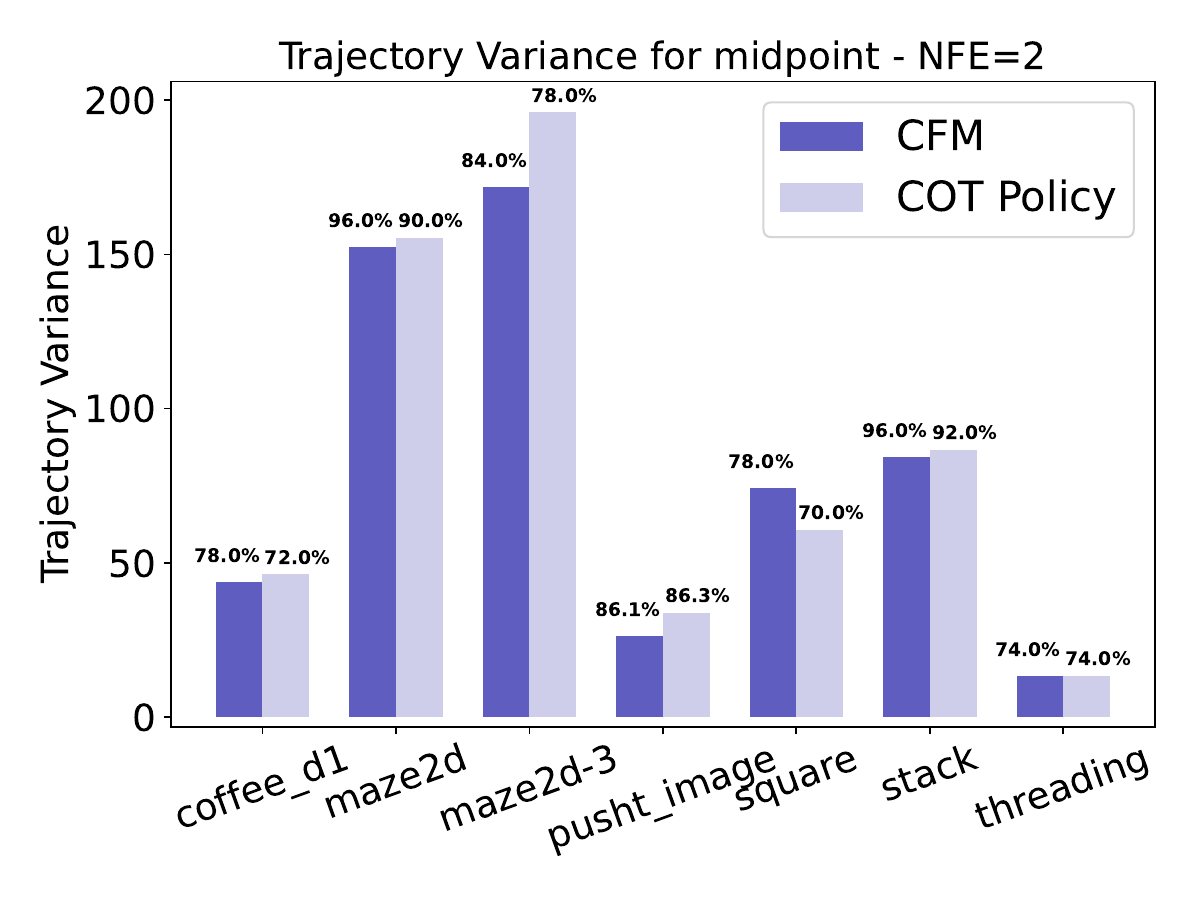}
    \end{subfigure}
    \hfill
    \begin{subfigure}{0.32\textwidth}
        \includegraphics[width=\linewidth]{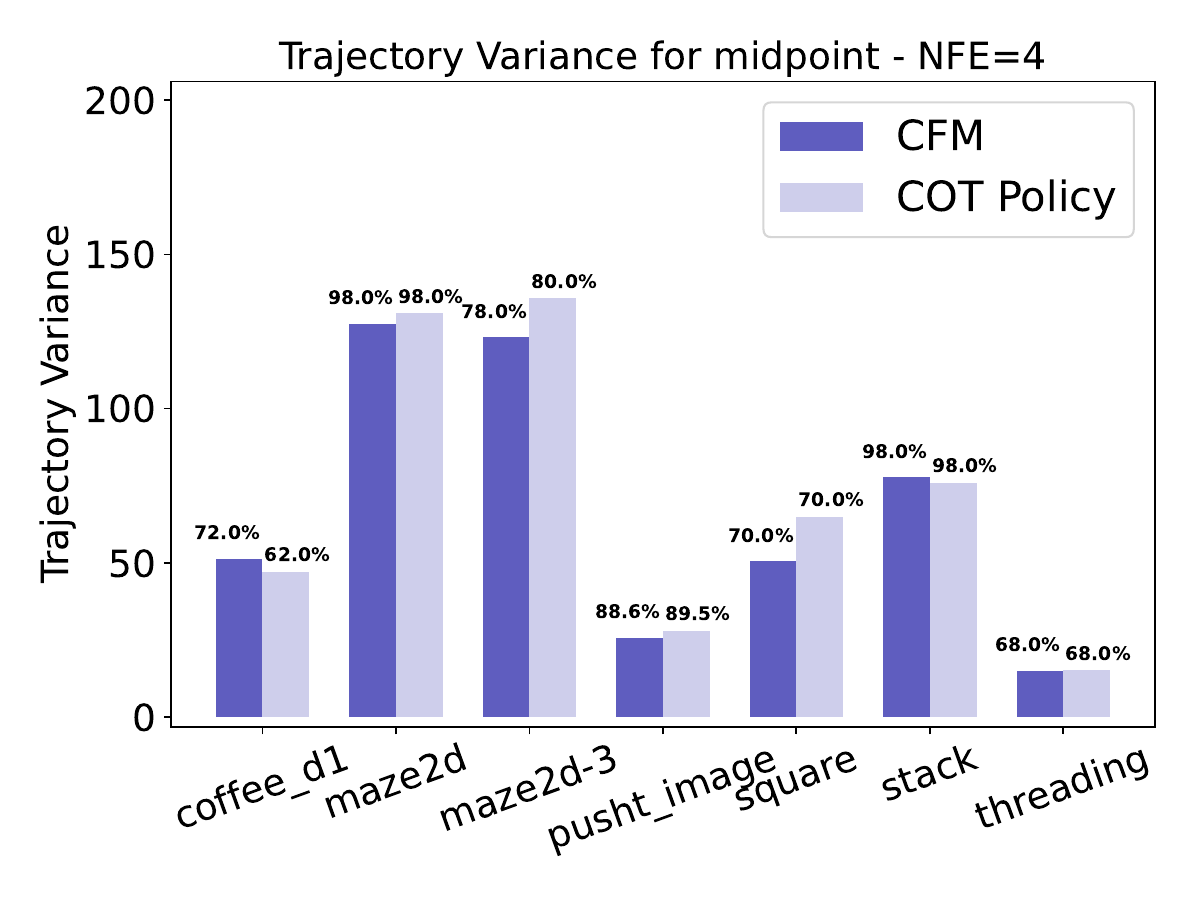}
    \end{subfigure}
    \hfill
    \begin{subfigure}{0.32\textwidth}
        \includegraphics[width=\linewidth]{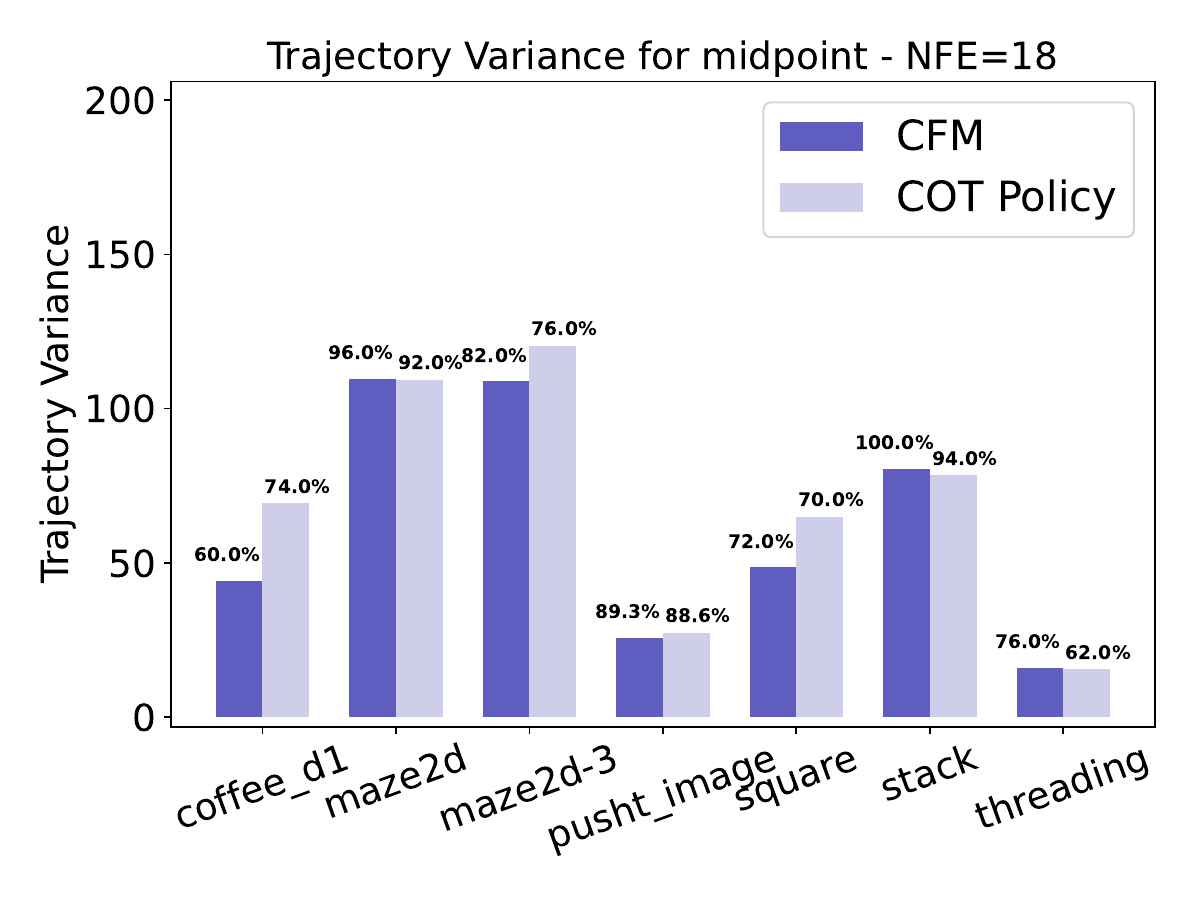}
    \end{subfigure}

    \caption{Trajectory variance comparison across six tasks using CFM and COT Policy \textbf{Top:} using the \texttt{euler} solver and \textbf{Bottom:} using the \texttt{midpoint} solver. Trajectory variance and success rates were computed over an average of 50 environemnt rollouts with random environment seeds (in contrast to \Cref{fig:cot_vs_cfm_multimodal} where enviornment seed was the same for all rollouts).}
    \label{fig:traj_variance_grid}
\end{figure}

\subsection{COT Policies without clustering}
\label{sec:appendix_cot_wo_clustering}
As we explain in \Cref{sec:number_of_clusters}, clustering alleviates the need for tuning $\gamma$ to a value that balances condition and action coupling. To demonstrate the need for fine-tuning $\gamma$ we report the performance of a variant of COT policy which uses \Cref{eq:gamma} for $\gamma$ and no clustering for the \texttt{push-t} and \texttt{coffee\_d1} tasks. As per \Cref{tab:cot_no_clustering}, clustering leads to increased success rates. Similar performance could potentially be achieved for COT without clustering, however this would require expensive tuning of $\gamma$ for every dataset. In contrast, clustering allows us to keep the number of clusters and $\gamma$ fixed throughout all tasks and still observe higher success rates and multimodal behaviors for low NFE.

\begin{table}[t!]
    \centering
    \resizebox{0.7\linewidth}{!}{
    \begin{tabular}{@{}l c|c c|c}
        \toprule
        \textbf{Method} & \textbf{NFE} & \texttt{push-t} & \texttt{coffee\_d1} & \textbf{Avg} \\
        \midrule
        \footnotesize COT Policy (w/o clustering) & 2 & 0.858 & 0.690 & 0.774 \\
        \footnotesize COT Policy (with clustering) & 2 & \textbf{0.878} & \textbf{0.787} & \textbf{0.833} \\
        \bottomrule
    \end{tabular}
    }
    \vskip 0.1in
    \caption{Performance comparison on \texttt{push-t} and \texttt{coffee\_d1} tasks.}
    \label{tab:cot_no_clustering}
\end{table}

\section{Hyperparameters of COT Policy}

\subsection{The effect of \texorpdfstring{$\gamma$}{gamma} on the minibatch OT solution}\label{sec:effect_of_gamma}

The hyperparameter $\gamma$, used in \eqref{eq:cond_cost}, regulates the effect the conditions $c_0, c_1$ have in the coupling process using unconditional minibatch OT. In order to retrieve an empirical coupling $\pi(x_0, x_1)$  that approximates the conditional OT coupling between $x_0$ and $x_1$, the second term of the cost \eqref{eq:cond_cost} needs to dominate (large $\gamma$) the first to prioritize pairing by condition. If the two terms have similar contributions to the cost (small $\gamma$), then neighboring noise $x_0$ and target samples $x_1$ might be coupled despite having different conditions. As already mentioned in \Cref{sec:conditional_ot_couplings}, when $\gamma=0$ we recover OT-CFM as noise and target samples are coupled solely based on their Euclidean distance.

\begin{figure}[t!]
\begin{center}
\centerline{\includegraphics[width=\columnwidth]{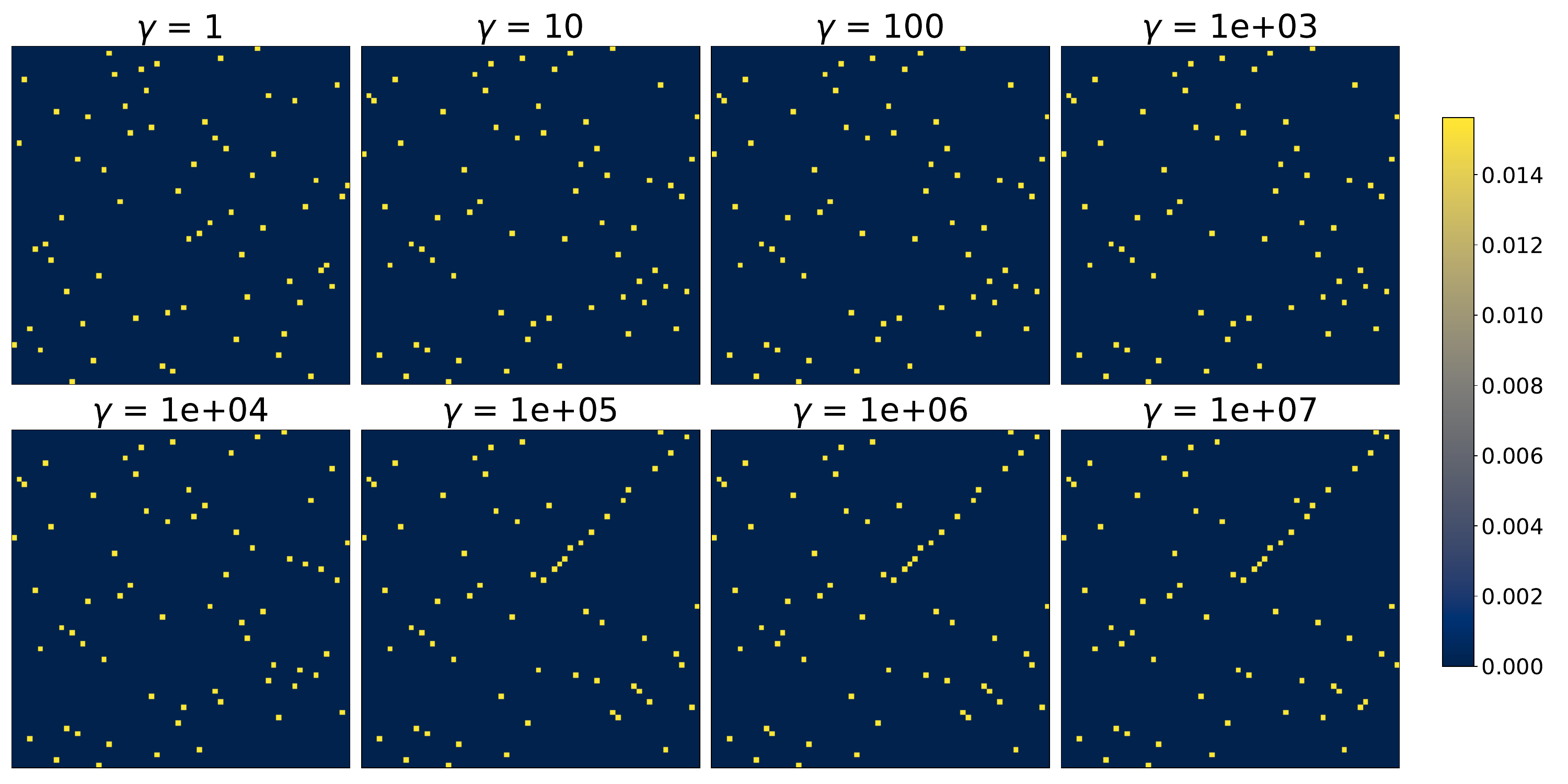}}
\caption{OT matrix for a range of condition scaling parameters $\gamma$. The OT matrix is consistent for medium to large values of $\gamma$ ($[10, 10000]$), but numerical errors are observed for larger values of $\gamma$.}
\label{fig:ot_maps}
\end{center}
\end{figure}

To get the desired empirical coupling that prioritizes coupling by condition, we therefore need to set $\gamma$ to a large value. However, it is possible that large values can lead to numerical instability. As can be seen by~\Cref{fig:ot_maps}, the OT matrix that we get from performing minibatch OT on the tuples $(x_0,c_0)$ and $(x_1, c_1)$ is consistent for $\gamma=\{10, 100, 1000, 10000\}$. For larger values ($\gamma>10000$) we observe that the diagonal dominates the top part of the OT matrix. This practically means that the $i$-th target point gets coupled with the $(|B_1|-i)$-th source point. This coupling is similar to the independent coupling ($i$-th target point get coupled with the $i$-th source point) and is the result of numerical errors encountered while solving the optimization problem of the Earth-Movers Distance (EMD) problem. 

To avoid behavior similar to (I)-CFM, while still prioritizing condition coupling, we set $\gamma$ for each batch based on the following formula:
\begin{align}
\label{eq:gamma}
    \gamma = 10 \cdot \frac{\overbrace{\sum_{i=0}^{|B_1|}\|x_0^{(i)}-x_1^{(i)}\|^2}^{\text{Avg. sample distance}}}{\underbrace{\sum_{j=0}^{|B_1|}\|c_0^{(j)}-c_1^{(j)}\|^2}_{\text{Avg. condition distance}}}
\end{align}
With this choice of $\gamma$ the second term of \eqref{eq:cond_cost} is scaled to be on average 10 times larger than the first. We have empirically found that choosing $\gamma$ based on \eqref{eq:gamma}, numerical errors in the minibatch OT calculations are avoided and in condition coupling is prioritized. 

\subsection{The effect of the number of clusters on COT Policy}
\label{sec:number_of_clusters}

Clustering the observations in COT Policy has a similar effect to setting $\gamma$ to a value that balances sample and condition distance. 
This is because samples $x_0^{(i)}, x_1^{(i)}$ from the same clusters ($c_0^{(i)}= c_1^{(i)}$) are coupled based on their Euclidean distance (according to \Cref{eq:cond_cost}), despite the fact that their corresponding actual observations might be different ($o_0^{(i)}\neq o_1^{(i)}$).
Finding a suitable $\gamma$ for every empirical distribution is not straightforward, but choosing a number of clusters is simpler based on the following observation:

\textbf{Proposition.} \textit{In tasks with continuous conditions, like visuomotor control, given a large enough $\gamma$, COT Policy approximates OT-CFM as $K\rightarrow0$ and (I)-CFM as $K\rightarrow|\mathcal{D|}$. }

Although we don't provide full proofs for the above proposition, we give some intuitions about the connection between CFM, OT-CFM, and COT-CFM in the extreme cases where $K=1$ and $K=|\mathcal{D}|$.

\paragraph{Clustering with $K=1$ cluster.} When we have a single cluster, then $c_0^{(i)}= c_1^{(i)},\ \forall i$. This reduces the cost of \Cref{eq:cond_cost} to:
\begin{align}
    \label{eq:unconditional_cost}
    c((x_0,c_0),(x_1,c_1))=\|x_0-x_1\|^2
\end{align}
As already explained in \Cref{sec:effect_of_gamma}, this renders COT-CFM (and consequently COT Policy) equivalent to OT-CFM.

\paragraph{Clustering with $K=|\mathcal{D}|$ clusters.} When the number of clusters is the same as the dataset size, then every observation $o^{(i)}$ acts as a separate cluster and therefore $c_1^{(i)}\neq c_1^{(j)},\ i\neq j$. The noise conditions in $B_0$ are random permutations of those in $B_1$ and therefore $c_0^{(i)}\neq c_0^{(j)},\ i\neq j$ and $c_0^{(i)}\neq c_1^{(j)},\ i,j\neq i, p(i)$, where $p(i)$ is the index of the sample of $B_1$ that was permuted to sample $i$ in $B_0$. This basically means that every noise condition has exactly one identical target condition. Assuming a large enough $\gamma$ that renders the second term of \Cref{eq:cond_cost} dominant, a coupling that would make it equal 0 would be the global optimum of the EMD optimization. That coupling exists and is constructed by the pairs $i, p(i),$ for $i=1, \dots, |B_1|$ since $c_0^{(i)}=c_1^{(p(i))}$. This is equivalent to a random permutation and therefore there is no dependency between the coupled samples. For this reason, COT-CFM with $K=|\mathcal{D}|$ is equivalent to (I)-CFM.

\end{document}